\documentclass[letterpaper, 10pt, conference]{IEEEtran}
\IEEEoverridecommandlockouts

\usepackage{siunitx}

\usepackage{float}

   \usepackage[pdftex]{graphicx}

  \graphicspath{{../pdf/}{../jpeg/}}
  \DeclareGraphicsExtensions{.pdf,.jpeg,.png}
  
  \usepackage{amssymb} 
  \usepackage{graphicx}

\usepackage[natural]{xcolor}

\usepackage{changes}

\usepackage[cmex10]{amsmath}
\usepackage{subcaption}
\usepackage{url}

\def\BibTeX{{\rm B\kern-.05em{\sc i\kern-.025em b}\kern-.08em
		T\kern-.1667em\lower.7ex\hbox{E}\kern-.125emX}}

\begin{document}

\title{\LARGE \bf 
Starting Movement Detection of Cyclists Using Smart Devices} 





\author{Maarten Bieshaar, Malte Depping, Jan Schneegans, and Bernhard Sick  \\
	\thanks{M. Bieshaar, M. Depping, J. Schneegans, and B. Sick 
		are with the \newline Intelligent Embedded Systems Lab, University of Kassel,
		Kassel, Germany \newline
		{\tt\footnotesize mbieshaar@uni-kassel.de, \newline
			mdepping@student.uni-kassel.de, jschneegans@student.uni-kassel.de, bsick@uni-kassel.de}}

}

\maketitle

\begin{abstract}
	

In near future, vulnerable road users (VRUs) such as cyclists and pedestrians will be equipped with smart devices and wearables which are capable to communicate with intelligent vehicles and other traffic participants. 
Road users are then able to cooperate on different levels, such as in cooperative intention detection for advanced VRU protection.
Smart devices can be used to detect intentions, e.g., an occluded cyclist intending to cross the road, to warn vehicles of VRUs, and prevent potential collisions. 
This article presents a human activity recognition approach to detect the starting movement of cyclists 
wearing smart devices. 
We propose a novel two-stage feature selection procedure using a score specialized for robust starting detection reducing the false positive detections and leading to understandable and interpretable features. 
The detection is modelled as a classification problem and realized by means of a machine learning classifier.
We introduce an auxiliary class, that models starting movements and allows to integrate early movement indicators, i.e., body part movements indicating future behaviour. In this way we improve the robustness and reduce the detection time of the classifier.  
Our empirical studies with real-world data originating from experiments which involve 49 test subjects and consists 
of 84 starting motions show that we are able to detect the starting movements early. Our approach reaches an $\pmb{F_1}$-score of 67\% 
within 0.33\,s after the first movement of the bicycle wheel.
Investigations concerning the device wearing location show that for devices worn in the trouser pocket the detector 
has less false detections and detects starting movements faster on average. 
We found that we can further improve the results when we train   
distinct classifiers for different wearing locations. In this case we reach an 
$\pmb{F_1}$-score of 94\% with a mean detection time of 0.34\,s for the device worn in the trouser pocket.
\end{abstract}



%
\IEEEpeerreviewmaketitle

\section{\large Introduction}
\label{sec_introduction}
\subsection{Motivation}

In our work, we envision future mixed traffic scenarios where  
automated cars, trucks, sensor-equipped infrastructure, 
and other road users equipped with 
smart devices or other wearables are interconnected by means of ad hoc networks. 
This allows the traffic participants to cooperate, i.e., determine and 
maintain local models of the surrounding traffic situations.
Vulnerable road users (VRUs) will still play an important role in future urban traffic. To avoid accidents, it is not only important to detect VRUs, but also to anticipate their intentions. Although modern vehicles are equipped with forward looking safety systems, dangerous situations for VRUs may still occur as a result of occlusions or sensor malfunctions.

\begin{figure}[t]
	\centering
	\includegraphics[width=0.45\textwidth, clip, trim= 20 30 40 20]{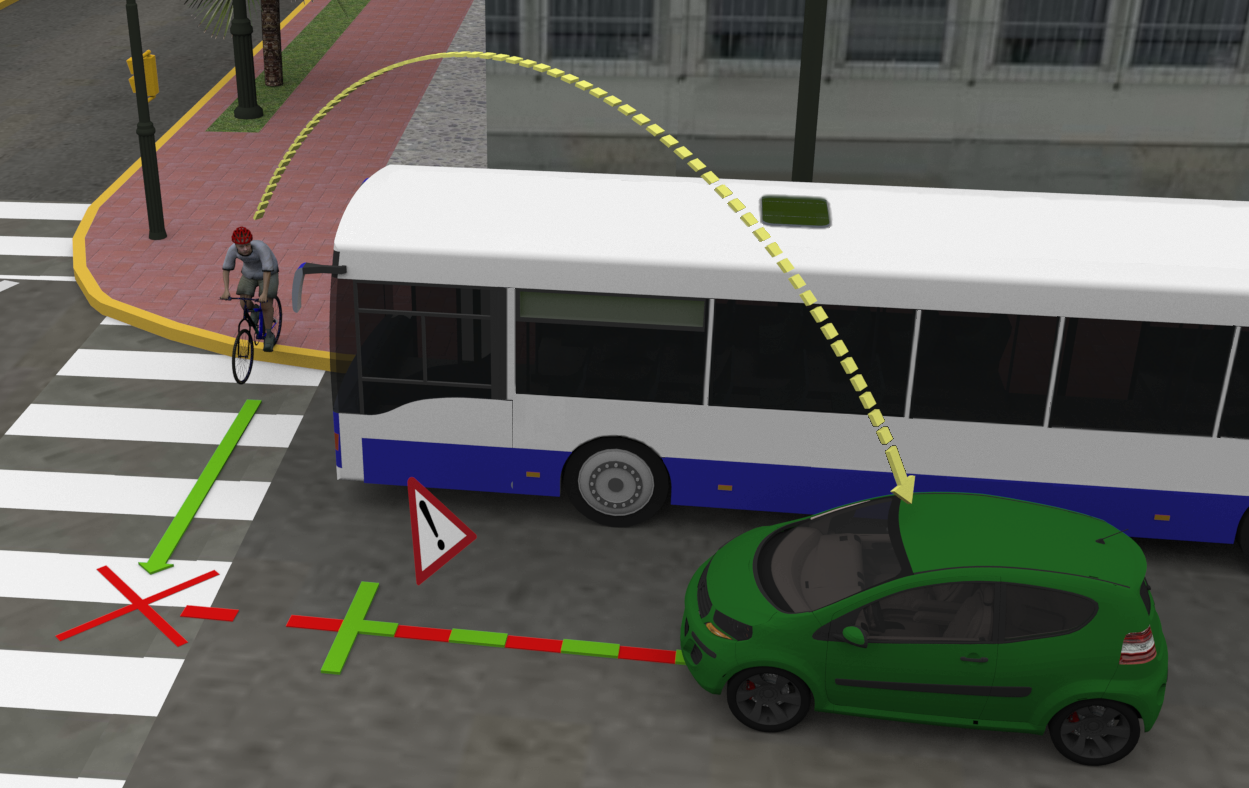}
	\vskip 4mm
	\caption{Typical dangerous situation in urban environment. The cyclist intending to cross the 
		road is occluded by the bus and cannot be seen by the approaching vehicle.
		The smart device worn by the cyclist anticipates the starting movement and transmits (indicated by dashed line) to the approaching vehicle such that a potential collision is avoided.
	}
	\vskip -6mm
	\label{fig:cyclist_crossing}
\end{figure}

Figure~\ref{fig:cyclist_crossing} sketches such a typical dangerous occlusion situation. A cyclist intends to cross the street while a vehicle hidden behind the bus is approaching. A smart device worn by the cyclist can anticipate the starting movement and communicate the detected intention to the approaching vehicle to warn the driver or initiate a braking maneuver. 

Smart devices and other wearables are capable of detecting the position using the device's 
integrated  global navigation satellite system (GNSS) 
and detecting movement transitions early using their inertial sensors.
Especially the latter are of great interest, since data from the inertial sensors 
is not affected by GNSS outage (a phenomena often encountered in urban areas) and 
it is available at high sampling rates. Hence, it is suitable to quickly detect transitions between movement states of VRUs, e.g., waiting and moving.
In this article, the focus is set on detecting cyclists' starting movement fast and yet reliable using inertial sensors. Our approach to detect starting movements is based on human activity recognition (HAR)~\cite{Bulling2014THA} and machine learning techniques.

The difficulty concerning smart device based starting movement detection is not whether we will detect a starting movement but to detect it as early as possible, i.e., in the range of a few hundred milliseconds after the actual starting movement. 
For illustration consider the following urban scenario: An automated vehicle is approaching with 50\,km/h and has a braking deceleration of 8\,m/$s^2$. If it initiates the braking 15\,m before a crossing cyclist, then it will stop 3\,m ahead of the cyclist. 
After the cyclist has entered the driving corridor, the automated system only has $0.58$\,s to initiate a braking maneuver and to prevent an accident. 
Nevertheless, the detector must also be robust, i.e., avoid false positive detections potentially leading to unnecessary emergency braking maneuvers.

%
%
%


\subsection{Main Contributions and Outline of this Paper}

Our main contribution is an approach based on HAR and machine learning using inertial sensors to detect cyclists' starting movements in real world traffic scenarios. We model the detection as a classification problem.
This article considers the following new aspects to detect cyclists' starting movements:


\begin{itemize}
	\item An improved starting movement detection process using 
	a novel training procedure based on an auxiliary class modelling the transition between waiting and moving. 
	It reduces the false positive detections and leads to earlier starting movement detections.
	\item A two-stage feature selection procedure to select robust features. It reduces false positive detections and leads to understandable and interpretable features. 
	\item 
	Investigation on different device wearing locations in which 
	we found that wearing the device in the trouser pocket leads to the best results. 
	\item Training of distinct classifiers for different wearing location to improve detection results.
\end{itemize}

The remainder of this article is structured as follows:
In Section~\ref{sec_related_work} related work is presented, 
 Section~\ref{sec_method_overview} describes our methodology for starting movement detection. The data acquisition and evaluation methodology is presented in Section~\ref{sec_evaluation} and the experimental results are reviewed in Section~\ref{sec_ResultsOutline}. Lastly, in Section~\ref{sec_conclusion}, the main conclusions and open issues for future work are reviewed.

%

\section{Related Work} \label{sec_related_work}
Many dangerous situations involving vehicles and VRUs occur in urban areas.
For better VRU protection, intelligent vehicles are equipped with active safety systems aiming to anticipate the VRU's intentions.  
We see the motion of a VRU as a sequence of activities or basic movements (e.g., standing or moving) 
and trajectory of certain body points (e.g., center of gravity or head) in 
the 3D space. Basic movement detection and trajectory forecasting are part of what we term as intention detection. 
In this article, we focus on detecting the starting movement. An early basic movement detection can support the trajectory forecast~\cite{mbBZD+17}.

Most of the conducted research concerning intention detection relies on computer vision based solution. 
In~\cite{KG14, Kooij2014}, the authors compare different approaches involving the past VRU positions and additional features derived from camera images. The approaches for pedestrian intention detection are based on recursive Bayesian filter (e.g., Kalman filter) and machine learning techniques, e.g., Gaussian process dynamical models, dynamic Bayesian networks, and probabilistic hierarchical trajectory matching. 
In~\cite{ZKS+18}, Zernetsch~et~al. use machine learning techniques, i.e., support-vector machines and  convolutional neural networks, to image sequences of an infrastructure-based camera system for early starting intention detection of cyclists. Their approach safely detects starting movements on average 0.14\,s after the cyclist starts moving. 
Although all of these techniques show promising detection results they all rely on camera- or laser 
scanner-based perception, which might not always be available (e.g., in case of occlusion). Cooperative 
approaches involving smart devices can alleviate this.

In~\cite{Liebner2013}, Liebner~et~al. present a bicycle warning system to warn turning vehicles
before approaching cyclists using the smartphone's integrated GPS and 3G HSDPA for smartphone-to-vehicle 
communication. The authors mainly focus on the evaluation of GPS accuracy. 
They assume that the cyclists move with constant velocity for predicting future cyclists' trajectory (for \SI{3}{\second}). This and additionally the communication delay limits the application of their approach.
The authors in \cite{MSN17} propose a cooperative system using smartphones and vehicles. GPS information originating from the smartphone is used to resolve occlusion situations and to enhance the perceptual horizon of vehicles. The authors solely focus on GPS and cooperative perception. They do not consider any other smartphone-based detectors.



A prototype system for cooperative safety applications involving a cyclist equipped with a smartphone 
is proposed by Thielen~et~al. in~\cite{Thielen2012}. The authors show a prototype application that warns a vehicle driver if the collision with a crossing cyclist is likely to occur within the next few seconds. The cyclist's trajectory forecast is based on GPS using a least squares fitting for extrapolation.
Due to the low GPS sample frequency and the extrapolation for prediction, fast movement transitions cannot 
be detected. In~\cite{Engel2013}, an approach involving Car2Pedestrian communication for pedestrian tracking is proposed.
It combines GPS data with inertial sensors allowing to transmit position and movement type to 
an approaching car. This allows to warn the pedestrian and the driver. Although their approach concerning 
pedestrian movement detection is similar to ours, they do not focus on the fast detection of movement 
transitions. Moreover, they focused on a single device wearing location using specialized features. 
In~\cite{BMD17}, the authors propose a smartphone-based approach to detect additional 
context information, e.g., detection of pedestrians leaving the curb, in order to improve the 
smart device based collision risk assessment.
In~\cite{mbMLT+17}, the authors present and compare different approaches to pedestrian path prediction for a time horizon of \SI{5}{\second} using a smartphone. They consider approaches based on artificial neural networks and dead reckoning. Their preliminary results on a small evaluation set consisting of only two pedestrians are promising. Nevertheless, their approach does not focus on detecting fast and critical movement changes nor do the authors evaluate different wearing locations. 

In our previous work \cite{mbBZD+17, decoint2BZH+18}, we presented a cooperative approach to 
cyclists' starting intention detection involving smart devices as an essential component. The approach presented in this article is an extension with special focus solely on smart devices.

\begin{figure*}[h]
	\centering
	\includegraphics[width=2.0\columnwidth, clip, trim=0 180 0 105]{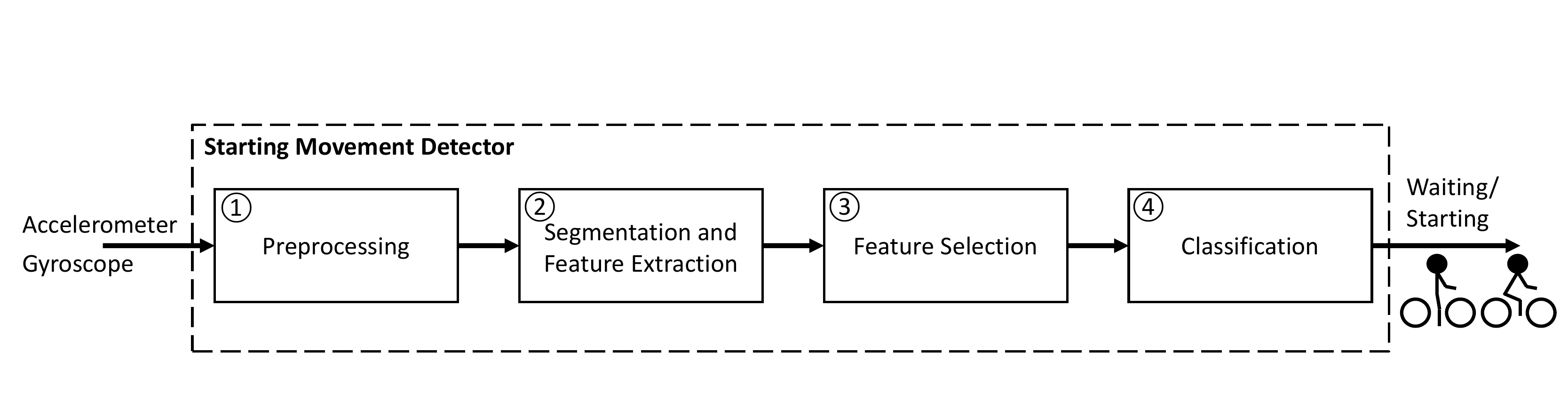}
	\caption{Process for starting movement detection based on smart devices. 
		It consists of the four stages: Preprocessing, segmentation and feature extraction, 
		feature selection, classification.
	}
	\label{fig:recognition_pipeline}
	\vskip -4mm
\end{figure*}

\section{\large Method}
\label{sec_method_overview}

Our approach aims to detect cyclists' movement transitions between waiting and moving (i.e., starting) as early as 
possible using smart devices carried by cyclists. The starting movement detector is realized by means of a HAR 
pipeline~\cite{Bulling2014THA}. 
A schematic of the starting movement detector consisting of four stages is depicted in Fig.~\ref{fig:recognition_pipeline}. 
The cyclists' starting movement detection is modelled as a classification problem, i.e., \textit{waiting}, \textit{moving}. 
Additionally, we consider an auxiliary class \textit{starting} modelling the transition between \textit{waiting} and \textit{moving}. The starting movement detector uses inertial data, i.e., accelerometer and gyroscope as input. The data is preprocessed (i.e., transformed in 
device attitude invariant representation), segmented, and features are extracted. Subsequently, to increase the generalization ability, feature selection is performed. Finally, the detection is realized by means of machine learning based classifiers, i.e., a support-vector machine with linear kernel (linear SVM) and an extreme gradient boosting classifier (XGBoost)~\cite{CG16}.

\subsection{Detection of Starting Movements}
\label{subsec:starting_movement_classifcation}

\begin{figure}[b]
	\centering
	\includegraphics[width=0.30\textwidth]{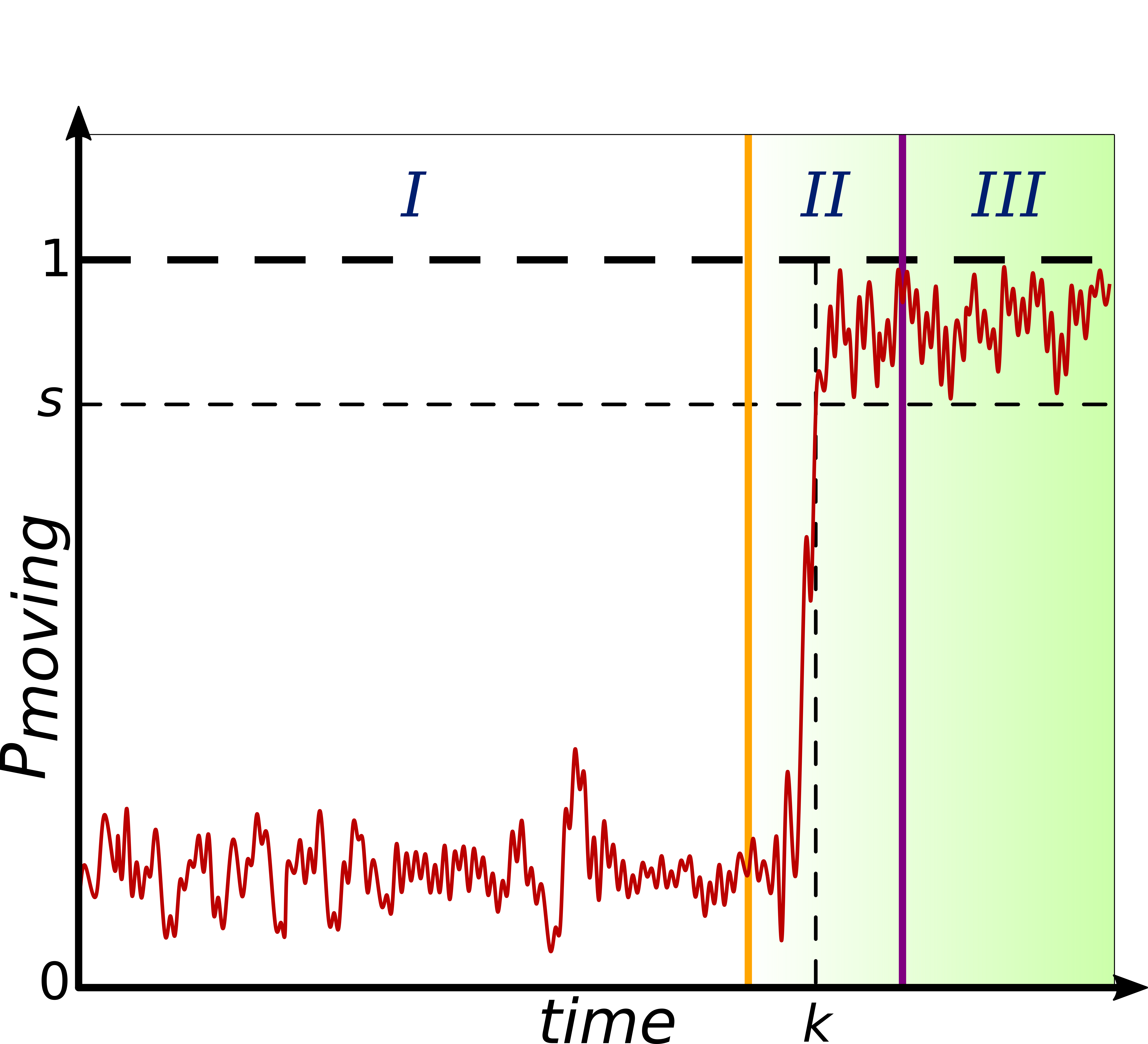}
	\vskip 3mm
	\caption{Exemplary detection output of one scene with the moving probability $P_{moving}$ (red), the labeled \textit{starting} time (orange), and the labeled \textit{moving} time (purple).}
	\vskip -5mm
	\label{fig:ex_net_out}
\end{figure}

In addition to the \textit{waiting} and \textit{moving} class, we introduce an auxiliary class \textit{starting}, which allows to integrate early movement indicators~\cite{mbHZD+17}, i.e., body movements happening before the beginning of \textit{moving}, indicating future behaviour. The \textit{starting} class is used in the model training and optimization process. 
It helps to avoid many false positive \textit{moving} detections, e.g., 
uncertain movements, which may either be classified as \textit{waiting} or \textit{moving} can now be classified as \text{starting}. 
Using the \textit{starting} class, we model the movement transition from \textit{waiting} to \textit{moving} as 
follows: $P_{waiting}$, $P_{starting}$, and $P_{moving}$ denote the probabilities assigned by the classifiers to the different classes. A phase is labelled as \textit{waiting} if neither the rear wheel of the bicycle is moving, nor is the cyclist performing a movement that leads to a starting movement. The time between the first visible movement of the cyclist that leads to a start and the first movement of the wheel of the bicycle is labeled as \textit{starting}. Finally, the sequence after the first movement of the bicycle wheel is labeled as \textit{moving}.
The \textit{starting} class is optional, since 
the labels are defined manually based on the evaluation of camera images.
A sample scene involving the three phases of the starting movement is depicted in Fig.~\ref{fig:ex_net_out}. 
The red line represents the probability $P_{moving}$ of the starting movement detection. 
In phase I this probability should be close to zero, during phase II the probability should increase, and reach a probability close to one in phase III. A detection is obtained by applying a threshold $s$ on  $P_{moving}$. The time of the first \textit{moving} classification is referred as starting 
detection time $t_d$.

A starting movement detector has to be robust, i.e., against false positive moving detections, and yet  
it has to be fast, i.e., low detection time. 
These are two opposing goals resulting in a trade-off which has to be solved.
How this trade-off is solved (i.e., which model parametrization is considered) 
depends on the rating of the goal. If the starting movement detection 
is used as supplementary information supporting the trajectory forecast~\cite{mbBZD+17}, then allowing 
a few false positive detections might be acceptable while in other cases having zero false positive 
detections is mandatory. 

%
%
%
%
\subsection{Preprocessing} \label{subsec:preprocessing}

\begin{figure}[b]
	\centering
	\includegraphics[width=0.5\textwidth, clip, trim=0 735 200 0]{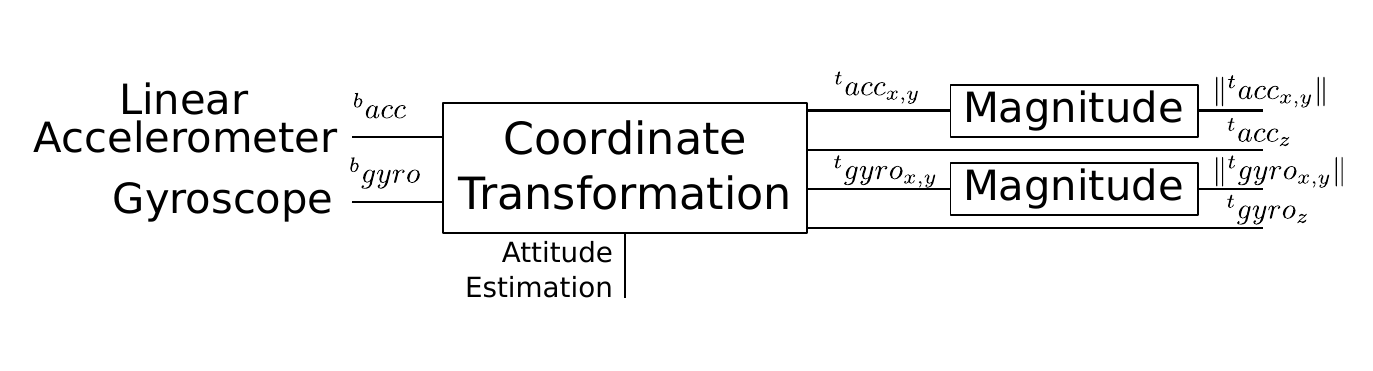}
	\caption{Preprocessing of linear accelerometer and gyroscope involving coordinate transformation and magnitude calculation.}
	\vskip -4mm
	\label{fig:preprocessing}
\end{figure}

The first stage of the starting movement detector is concerned with the sensor preprocessing.
Our approach uses the accelerometer and gyroscope sensor, which are sampled 
with a frequency of \SI{50}{\Hz}. The angular velocity measured by the gyroscope 
can be used to detect rotation movements, such as pedalling, while the accelerometer 
is better suited to detect linear movements, e.g., forward movement.
We use gravity compensated accelerometer data, referred as linear acceleration data. 
The gyroscope values are bias corrected using the calibration supplied by the manufacturer. 
A further drift compensation is not performed. Modern smart devices run sensor fusion algorithms. These incorporate data originating from gyroscope, accelerometer, and magnetometer to estimate the smart device orientation with respect to a global reference frame. 
The preprocessing described in the following is depicted in Fig.~\ref{fig:preprocessing}. 
The three components ($x$, $y$, and $z$) of the linear accelerometer ${}^{b}acc$ and gyroscope ${}^{b}gyro$ are transformed using the orientation estimation supplied by the smart device's operating system. The smart device coordinate system $b$ is referred to as body frame.
The linear acceleration and gyroscope data is now in a coordinate frame which is leveled with 
the local earth ground plane, i.e., the $z$-axis is pointing towards the sky. This coordinate 
frame is referred to as the local tangential frame $t$. 
The compass is not considered due to its sensitivity to a precise calibration~\cite{MGF+17} and possible magnetic perturbations. For this reason, we do not consider the precise transformation from this local frame to a global reference frame (e.g., north-east direction frame) or the difficult estimation of 
the device's orientation with respect to the VRU.
We resolve this issue by considering the magnitude of the linear accelerometer $|| {}^{t}acc_{x,y} ||$ and the gyroscope $|| {}^{t}gyro_{x,y} ||$ in the local tangential horizontal $x-y$ plane (also referred as ground plane). This representation is invariant concerning orientation. Hence, we avoid the challenging estimation of the transformation between the device orientation with respect to the VRU~\cite{MGF+17}. In addition, we consider the projection of the linear accelerometer ${}^{t}acc_{z}$ and the gyroscope  ${}^{t}gyro_{z}$ on the local vertical $z$-axis, i.e., the gravity axis. 

We do not consider the smart device integrated GPS for the following two reasons. 
First, GPS is not always available or noisy due to multipath effects especially in urban 
areas. Second, the sampling frequencies of \SI{1}{\Hz} provided by modern smart devices is too low to detect fast changes in cyclists' movements.

\subsection{Segmentation and Feature Extraction} \label{subsec:segmentation_feature_extraction}

In the second stage, we perform a sliding window segmentation using different window sizes of the transformed 
signals and consecutively extract features. Here, we consider features commonly used in HAR~\cite{Bulling2014THA}, such as the energy, minimum, and maximum, which are computed for 
sliding window sizes \SI{0.1}{\second}, \SI{0.5}{\second}, \SI{1.0}{\second}, and \SI{2.0}{\second}.
Features computed for different window sizes allow the detector to  
handle different time scales, i.e., features computed with the smaller window sizes capture short term dependencies and larger window sizes capture dependencies on a longer timescale.
Furthermore, we consider features based on orthogonal polynomial approximation up to the $3^{\mathrm{rd}}$ degree
for window lengths of \SI{0.5}{\second}, \SI{1.0}{\second}, and \SI{2}{\second}. These polynomial coefficients 
are in a least squares sense best estimators of the mean, slope, and curvature of the input signal~\cite{Fuchs2010}.
In addition, we consider the magnitude of the discrete Fourier transform (DFT) coefficients 
for window sizes \SI{0.64}{\second} and \SI{5.12}{\second}, 
as successfully applied for human walking speed estimation in~\cite{PPC+12}.
We only consider coefficients up to the $5^{\mathrm{th}}$ order since typically human motion 
is best represented by the lower frequencies.
In order to make the DFT coefficients independent of the energy, we normalize the coefficients with respect to the 
square root of the signal's overall energy within the respective window.
In total, $164$ features are computed.

\subsection{Feature Selection} \label{subsec:feature_selection}

\begin{figure}[h]
	\centering
	\includegraphics[width=0.48\textwidth]{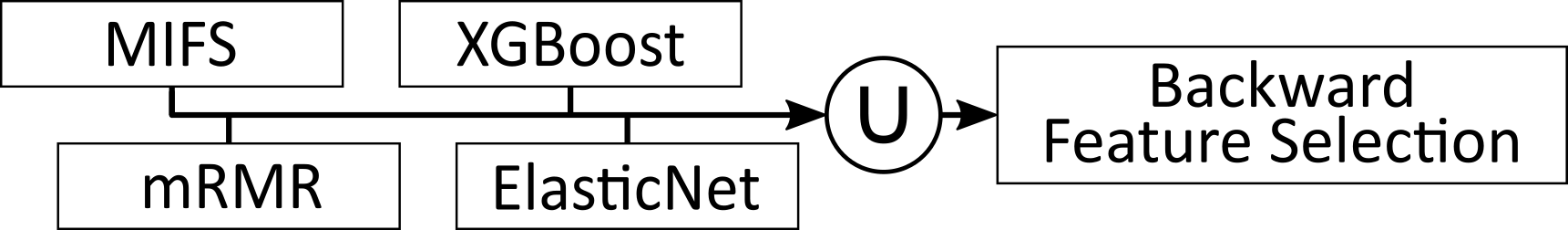}
	\vskip 4mm
	\caption{Two-stage feature selection approach: First we apply four different filters (i.e., MIFS, mRMR XGBoost, and elastic net) and union ten best scoring features of each filter, Second, we apply backward feature selection.}
	\vskip -4mm
	\label{fig:feature_selection}
\end{figure}

Selecting good features is the key for robust and yet fast detection.
The feature selection is realized in the third stage of the starting movement detector.
As with the final classification, we have to face the problem of highly
imbalanced classes. In order to compensate for this, we randomly  
undersample the \textit{waiting} class up to the size of the \textit{moving} class. 
Subsequently, we randomly oversample the \textit{starting} class, such that 
all three classes have equally many samples.
For the feature selection, we adopt the following approach: 
First, we apply filters~\cite{LCW+17} and then we fine-tune the feature selection 
by means of a wrapper approach. A schematic of this is depicted 
in Fig.~\ref{fig:feature_selection}. The goal of using filters is to 
pre-select a set of potentially meaningful features and to 
reduce the computational complexity. In order to get a large diversity 
concerning pre-selected features, we apply different filters, i.e., mutual information (MIFS), 
minimum redundancy maximum relevance (mRMR) feature selection and two model-based selection techniques based on elastic net and XGBoost.
MIFS selects features which generally have a high mutual information score for the 
classification target, whereas mRMR selects features with high mutual information which additionally do not overlap much. 
The model-based techniques select features suitable for the respective classifier, i.e., the XGBoost and linear discriminative models.

We consider the union of the ten best scoring features of each filter resulting in at most $40$ different features.
Then, in the second stage of the feature selection, 
we use these previously selected features in a wrapper approach in 
combination with the currently considered detector. This fine-tunes the 
feature selection by only selecting those which are relevant for the 
optimization criteria. 
We use an $F_1$-score defined over starting scenes (described in Section~\ref{subsec_evaluation}) as optimization criteria to 
perform a backward feature selection~\cite{GE03} with the respective 
classifier. The parameters of both classifiers are fixed during feature selection.

\subsection{Classification}

In the fourth stage of the starting movement detector, 
the starting detection is realized by means of a frame-based classification using 
a linear SVM and XGBoost~\cite{CG16}. The frame-based classification is performed at discrete points with a frequency of \SI{50}{\Hz}. The linear SVM is chosen since it has proven a good generalization ability and 
its linear decision boundary can be efficiently evaluated when applying the classifier in the field. 
Especially the latter is appealing for the given evaluation frequency of \SI{50}{\Hz}.
The XGBoost algorithms has won many awards in current machine learning challenges and can be considered 
state-of-the-art. Nevertheless, the algorithm is computationally more involved.

The linear SVM was optimized using the Hinge loss. Moreover, due to the large number of 
training samples, we trained the linear SVM in the primal space. 
The XGBoost algorithm is an ensemble method based on classification 
and regression trees. We considered the logistic loss as the objective function underlying the 
XGBoost training with a regularization term controlling the model complexity (see~\cite{CG16} for more details). In order to further improve the generalization ability, we considered  regularization in form of learning rate shrinkage and random feature subsampling.

The classifier is trained on labeled data consisting of three classes, i.e., \textit{waiting}, \textit{starting}, and \textit{moving}. During run-time only the \textit{waiting} and \textit{moving} classes are considered for starting movement detection. 
As before for the feature selection, we randomly undersample the \textit{waiting} class 
and oversample the \textit{starting} class. The class ratio, i.e, fraction of samples of each class used for training, is an important factor influencing the design of the resulting starting movement detector. For example, 
putting strong emphasis on the \textit{waiting} class results in robust starting movement detection but high detection times. 
Instead of directly using the classification returned by the frame-based classifier, 
we favour to use class probabilities representing confidence estimates about the classification, i.e., detection.
This probability estimate can be used to design the starting detector, 
such that it reacts only upon exceeding a certain threshold applied on the \textit{moving} class probability.
Since neither the linear SVM nor the XGBoost classifier return proper probability estimates 
(e.g., the XGBoost tends to predict  overconfident probabilities), a probability calibration fitting an additional 
sigmoid (i.e., Platt calibration~\cite{Niculescu-Mizil2005PGP}) is performed. 

Before determining the final decision using the threshold on the \textit{moving} class probability, 
an additional smoothing is performed to reduce the amount of false positive detections.
This is realized by means of a soft voting ensemble approach 
i.e., running average of the \textit{moving} class probability.
\section{Data Acquisition and Evaluation}
\label{sec_evaluation}

\begin{figure}
	\centering
	\includegraphics[width=0.45\textwidth]{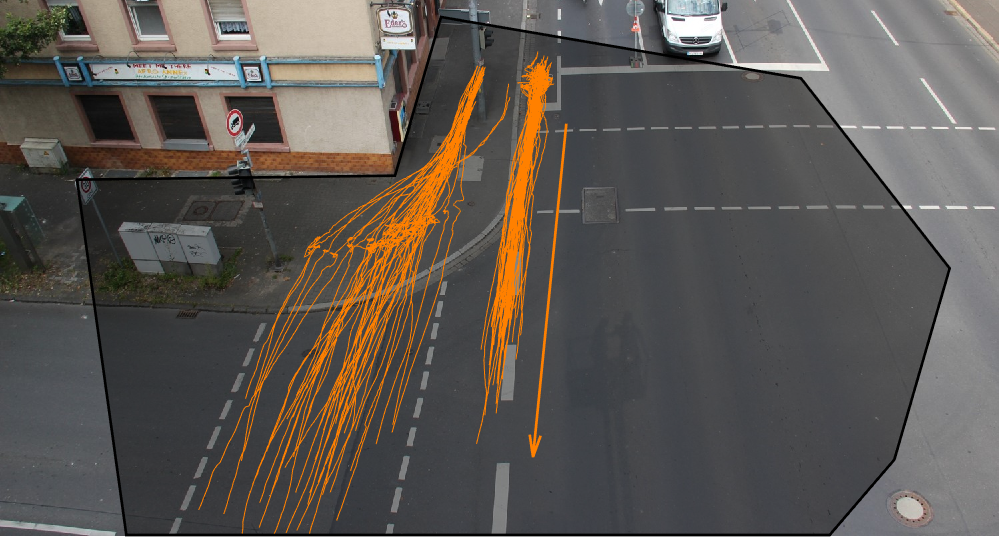}
	\vskip 4mm
	\caption{Overview of the intersection with all starting movements of instructed cyclists. The arrow is pointing into the starting direction.}
	\label{fig:trajectories}
	\vskip -4mm
\end{figure}

\subsection{Data Acquisition} \label{subsec_data_acquisition}

For the evaluation of our approach, we used a dataset consisting of cyclists's starting movements.
It contains 49 female and male test subjects. They were instructed to move between certain points at an intersection with public, uninstructed traffic while adhering to the traffic rules. Since there are two traffic lights at the intersection, we obtained 84 starting motions, with a maximum of two starting motions per test subject. 
The trajectories of the cyclists are plotted in Fig. \ref{fig:trajectories}. 
In order to label the starting movements as described in Section~\ref{subsec:starting_movement_classifcation}, 
we recorded the images of two HD cameras, which were installed at the 
intersection as part of a wide angle stereo camera system.
All test subjects were equipped with four smart devices worn at different body locations, i.e., wearing locations.
The smart devices used for evaluation are Samsung Galaxy S6 smartphones.
In our opinion these four locations listed in the following 
are typical representatives of wearing locations observed in everyday life.
The test subjects were equipped with a smartphone in their front trouser pocket.
Moreover, the test subjects were equipped with a smartphone placed in a backpack. 
Another smart device was placed in the front pocket of the test subjects' jackets at the height of their chests. 
For those test subjects which did not posses a front pocket at the chest  
the smartphone was mounted within a pocket located at the chest belt of the backpack. 
The fourth device was mounted at the rack of the bicycle. If no rack was available, then it was put into 
a bag mounted beneath the saddle. The motivation behind choosing the placement on the rack is that many cyclist 
carry their smartphone in a bag on their rack. 
 The smart devices were all placed in predefined orientations, e.g., for front trouser pocket: Upright position and 
 display facing outwards. This makes the experimental setup comprehensible and reproducible.  This does not limit the general applicability of the presented approach since it uses features computed on a device orientation invariant representation of the inertial data (cf. Section~\ref{subsec:preprocessing}).
We used different bicycles during the experiments, ranging from mountain and touring over city to racing bikes.
The dataset used in this article for evaluation of our approach is made publicly available\footnote{\url{https://git.ies.uni-kassel.de/intention_detection/starting_detection}}.

\subsection{Evaluation} \label{subsec_evaluation}
We performed the evaluation of our starting movement detection approach off-line using a ten-fold cross-validation over the VRUs. 

For detection performance assessment, we propose a scene-wise evaluation.
A scene comprises the timespan after stopping till the cyclist is leaving the field of view of the camera 
used for labeling. It only consist of a single \textit{waiting}, \textit{starting}, and \textit{moving} phase (cf.~Fig.~\ref{fig:ex_net_out}), 
whereby the \textit{starting} is optional, since some cyclists start moving right away without showing 
any early movements.  A scene is rated as false positive if the detection time falls into phase I.
If the detection is in phase II or  III, then it is rated as true positive. 
If the \textit{moving} class is not detected, then it is rated as false negative. 
Since every waiting phase ends in a starting phase, we do not consider true negatives.
The overall quality regarding robustness (i.e., avoiding false detection) 
of the detectors is evaluated by means of the $F_1$-score calculated over all scenes. 
For evaluation we additionally remove \SI{3}{\second} at the beginning of each waiting phase  
since we focus on detecting \textit{waiting} to \textit{moving} transitions, i.e., starting movements.

Besides the robustness, also the detection time $t_d$ is crucial. 
Therefore, we calculate the mean time difference $\overline{\delta_t}$ between the detection time $t_d$ and the start time of phase III $t_{III}$ of all true positives over all $L$  scenes (Eq. \ref{eq:mean_detection_time}).
Smaller values are better, even negative values are possible.

\begin{equation}
\overline{\delta_t}=\frac{1}{L}\sum_{i=1}^{L}(t_{di}-t_{IIIi})
\label{eq:mean_detection_time}
\end{equation}

The trade-off regarding robustness and a fast detection can be considered a multi-objective 
optimization problem~\cite{mie99}. For evaluation we adopt the concept of Pareto optimality. A solution is Pareto optimal 
if none of the involved objective functions can be further improved without deteriorating some of the other 
involved objectives. The set of Pareto optimal solutions is referred as Pareto frontier. Those solutions 
which are not Pareto optimal are referred as dominated solutions.
Without any additional weighting (i.e., rating which goal is more important) all Pareto optimal solutions are considered equally good. We create the Pareto frontiers by means of a parameter sweep, i.e., we evaluate the $F_1$-score and mean detection time for a set of different parameter configurations. Subsequently, we can calculate the Pareto frontier for this set.

\section{Experimental results}
\label{sec_ResultsOutline}
\subsection{Wearing Location} \label{sec:eval_wearing_location}

\begin{figure*}[ht!]
	\centering
	\vskip -4mm
	\begin{subfigure}[t]{0.49\linewidth}
		\centering
		\includegraphics[width=0.9\columnwidth]{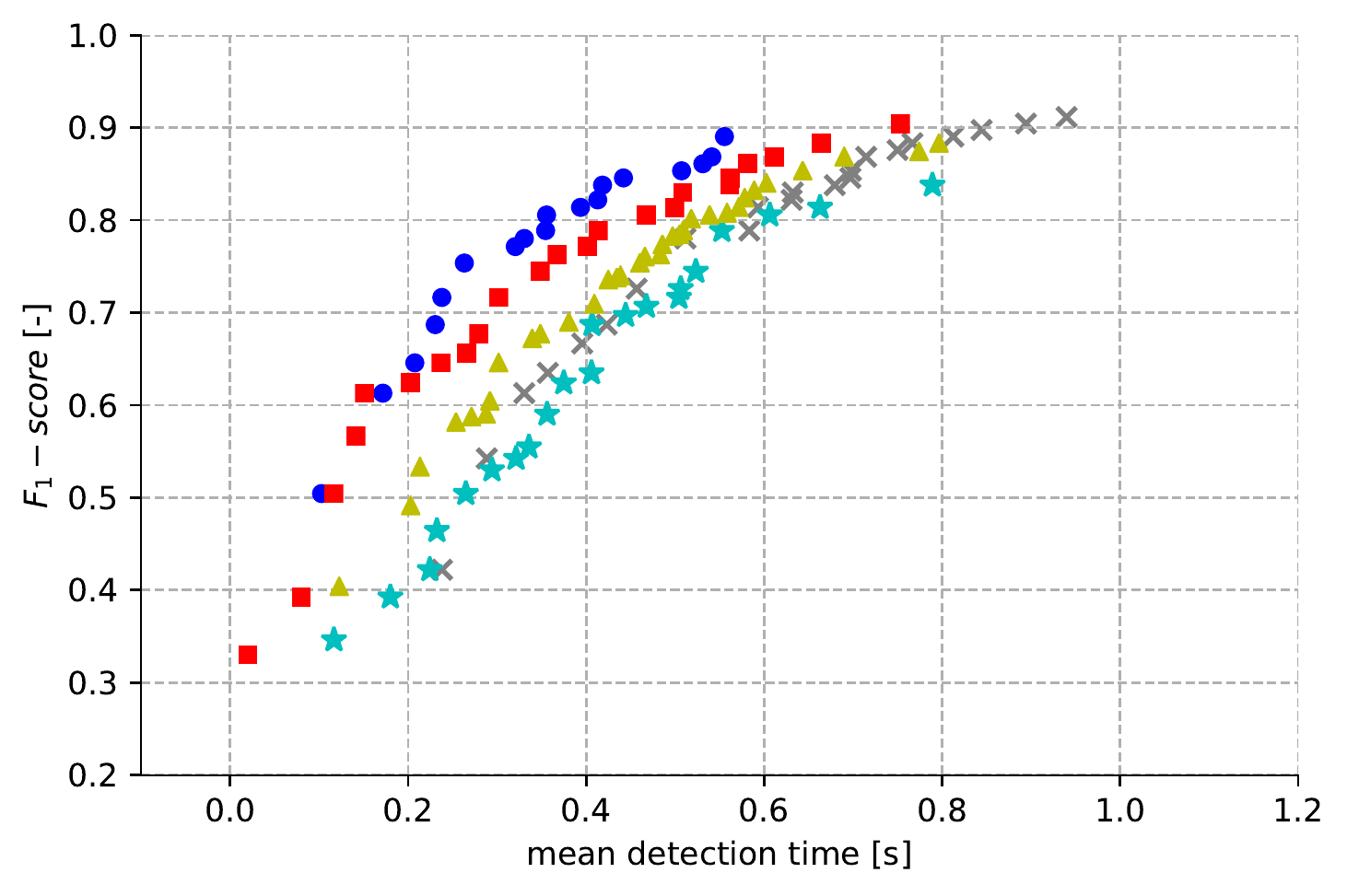}
		\vskip -2mm
		\caption{Pareto frontiers of \textit{location-agnostic} XGBoost.}
		\label{fig:xgb_all_pareto_frontier}
	\end{subfigure}
	\begin{subfigure}[t]{0.49\linewidth}
		\centering
		\includegraphics[width=0.85\columnwidth]{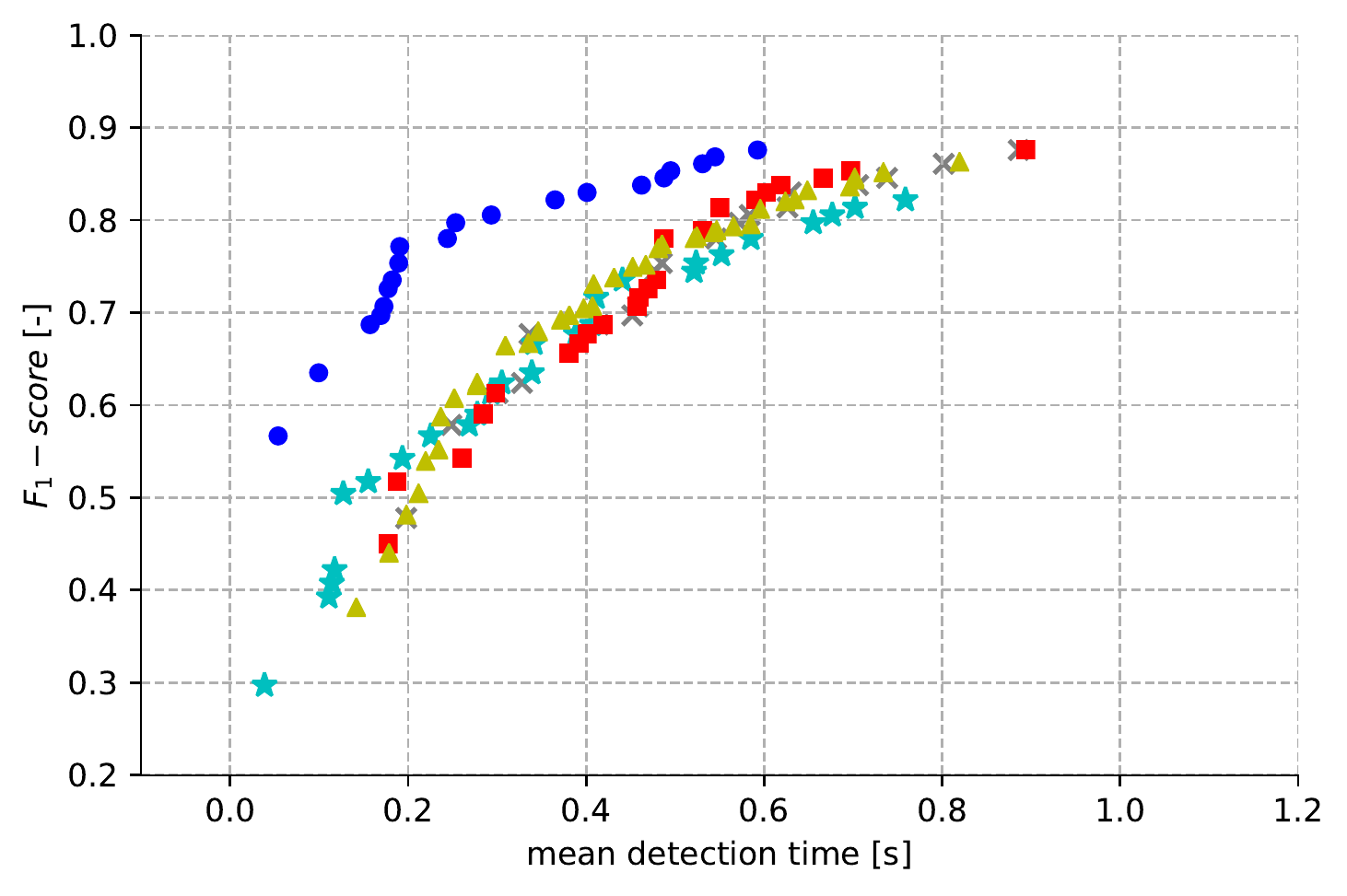}
		\vskip -2mm
		\caption{Pareto frontiers of \textit{location-agnostic} linear SVM.}
		\label{fig:lin_svm_all_pareto_frontier}
	\end{subfigure}
	\vskip 4mm
	\begin{subfigure}[t]{0.49\linewidth}
		\centering
		\includegraphics[width=0.9\columnwidth]{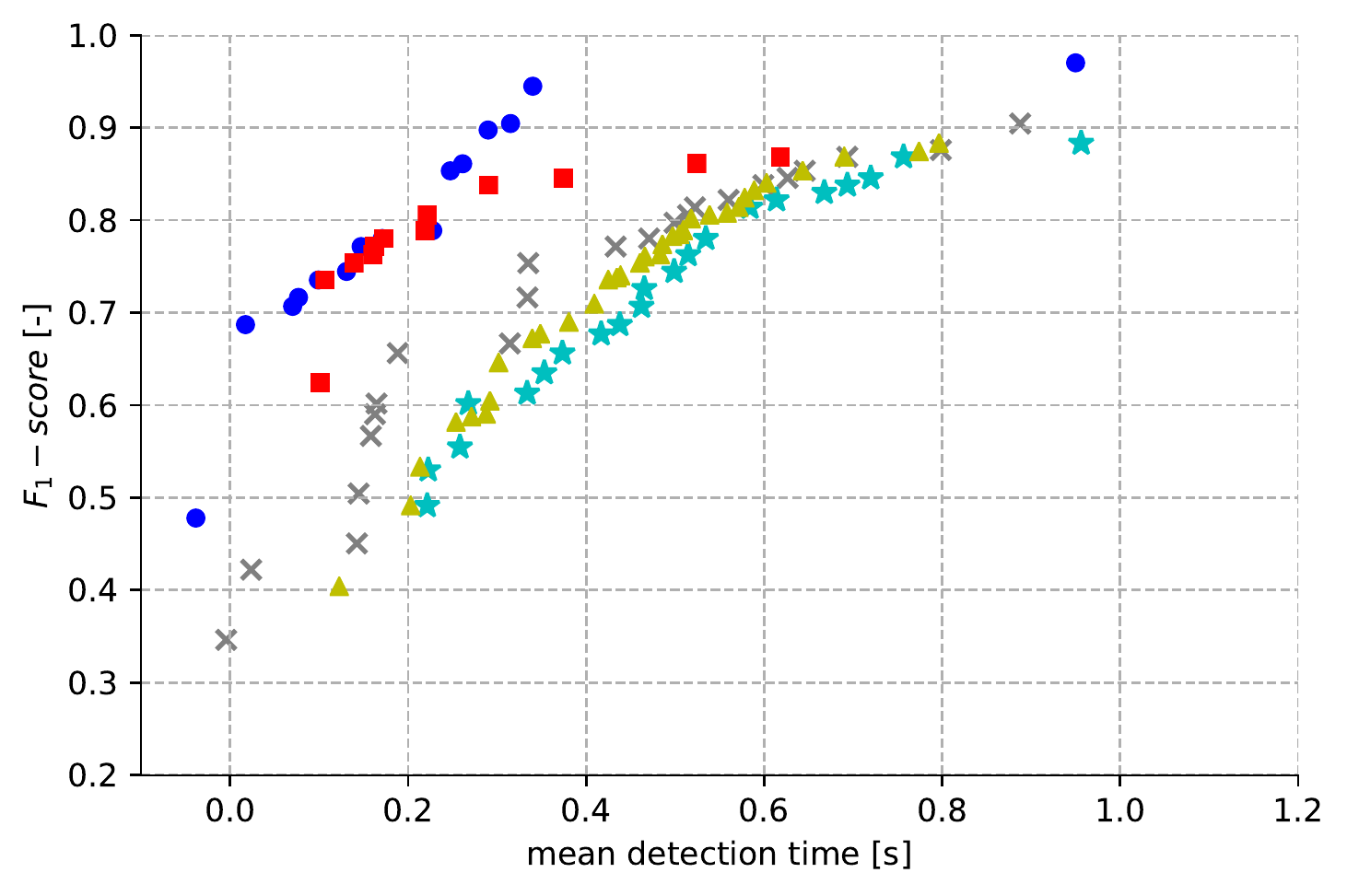}
		\vskip -2mm
		\caption{Pareto frontiers of \textit{location-specific} XGBoost.}
		\label{fig:xgb_pareto_frontier}
	\end{subfigure}
	\begin{subfigure}[t]{0.49\linewidth}
		\centering
		\includegraphics[width=0.85\columnwidth]{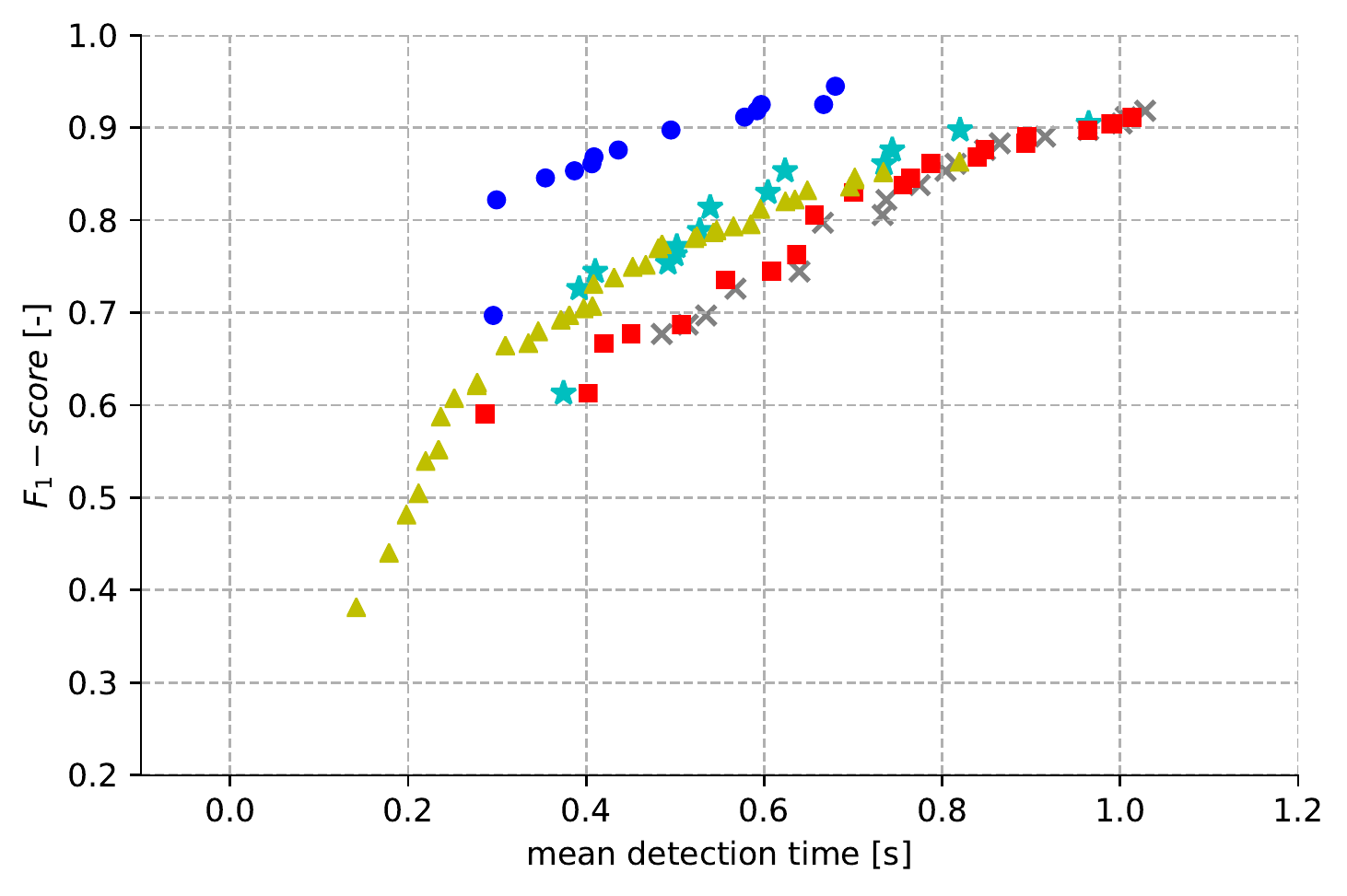}
		\vskip -2mm
		\caption{Pareto frontiers of \textit{location-specific} linear SVM.}
		\label{fig:lin_svm_pareto_frontier}
	\end{subfigure}
		\vskip 4mm
	\begin{subfigure}[t]{0.49\linewidth}
		\centering
		\includegraphics[width=0.9\columnwidth]{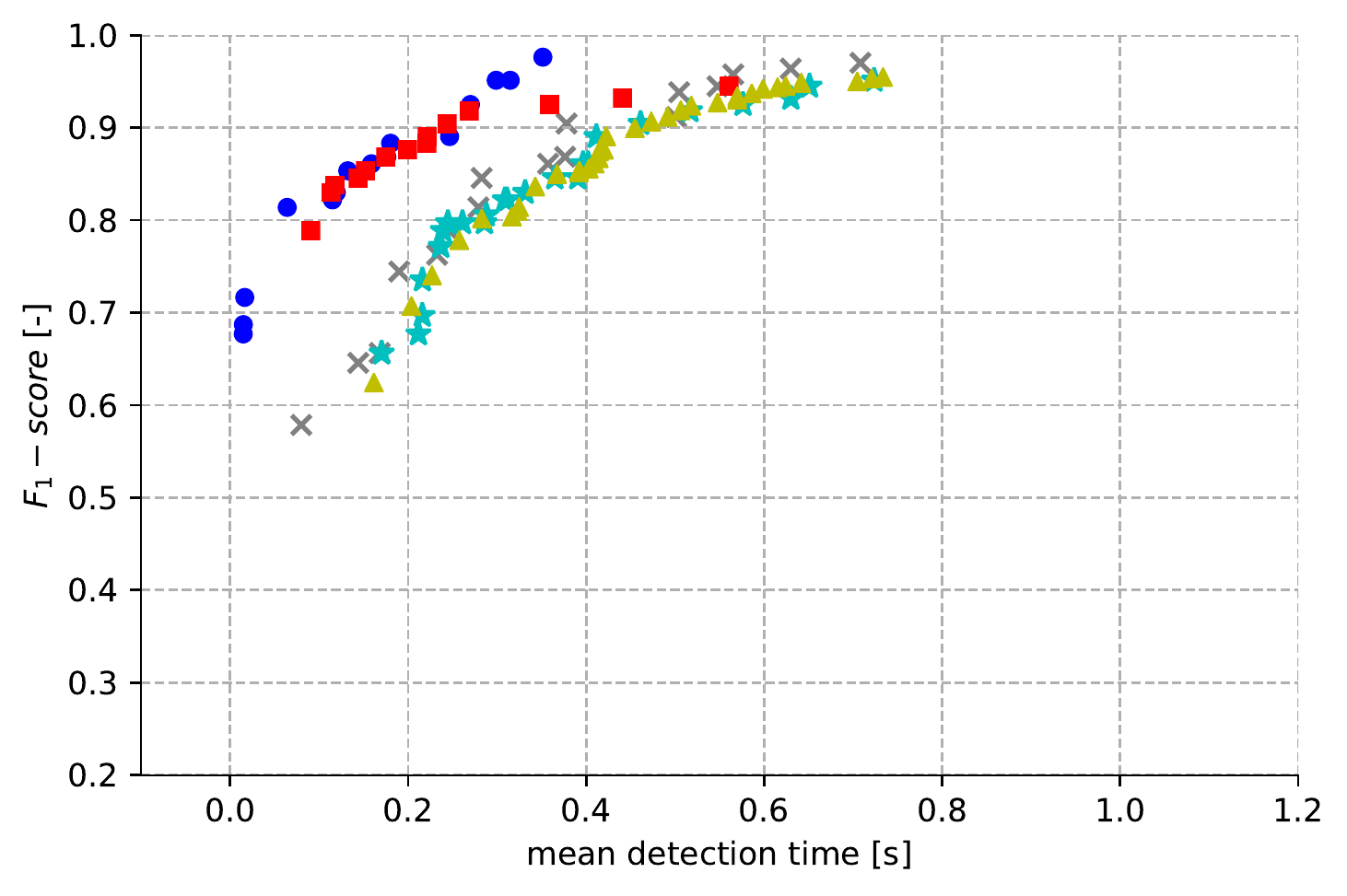}
		\vskip -2mm
		\caption{Pareto frontiers of \textit{location-specific} XGBoost for shortened starting scenes.}
		\label{fig:short_xgb_all_pareto_frontier}
	\end{subfigure}
	\begin{subfigure}[t]{0.49\linewidth}
		\centering
		\includegraphics[width=0.9\columnwidth]{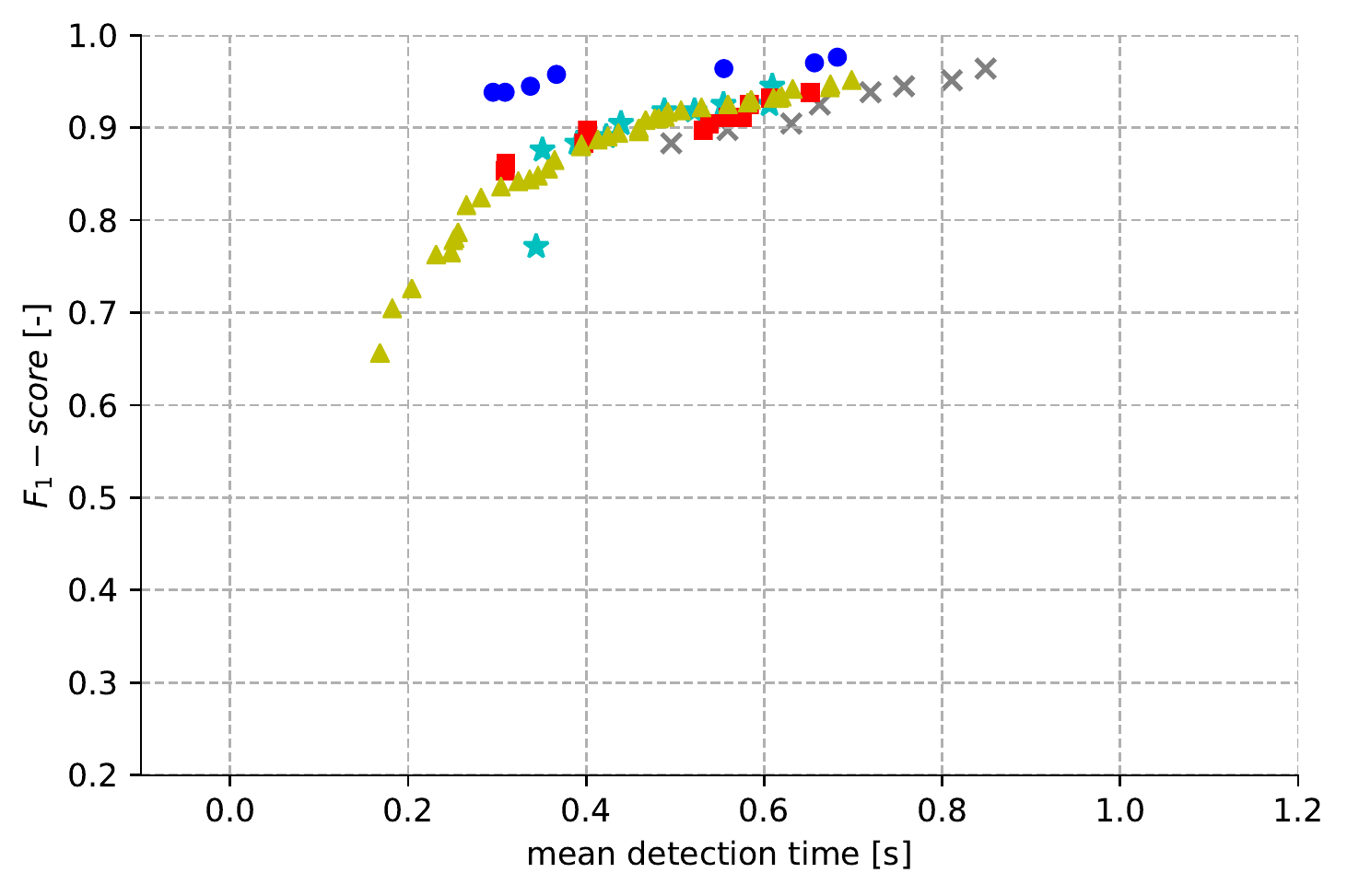}
		\vskip -2mm
		\caption{Pareto frontiers of \textit{location-specific} linear SVM for shortened starting scenes.}
		\label{fig:short_lin_svm_all_pareto_frontier}
	\end{subfigure}
	\vskip 5mm
	\caption{Pareto frontiers of $F_1$-scores and mean detection times for XGBoost and linear SVM classifiers 
		evaluated for different wearing locations (Blue dot: trouser pocket, Red square: backpack, Cyan star: jacket pocket/ chest, 
		Gray cross: Bicycle rack, and Yellow triangle: Average over all wearing locations using the respective \textit{location-agnostic} classifier).
		The two upper figures show the evaluation \textit{location-agnostic} classifiers and the two in the middle show the evaluation of the \textit{location-specific} classifiers. The lower two show the evaluation of the \textit{location-specific} classifiers on the shortened starting scenes.
	}
	\label{fig:res_wearing_location}
	\vskip -2mm
\end{figure*}

In this section, we evaluate and compare different models and parametrizations to detect cyclist starting movements.   
Moreover, we evaluate the detection performance with respect to four different wearing locations.
We trained separate linear SVM and XGBoost classifiers on data originating from devices worn at the four different 
wearing locations. These are referred as \textit{location-specific} classifiers. Additionally, we trained 
classifiers incorporating data from all four wearing locations. These are referred to as \textit{location-agnostic} classifiers.  

To create the Pareto frontier, we randomly sampled and evaluated 250 different parameter combinations for each \textit{location-specific} and \textit{location-agnostic} classifier. For the XGBoost classifier we considered  
the number of trees (50,~100,~200,~300,~500,~and~700), the maximum tree depth (between 3 and 10), 
and the learning rate (between 0.01 and 0.2) as parameters. For the linear SVM we only considered the penalty term (between $2^{-8}$ and $2^{8}$). 
In order to speed up the evaluation, we performed a random subsampling. 
We experimentally determined that at least approximately 7500 (30000) samples are required to achieve reasonable results
for a \textit{location-specific} (\textit{location-agnostic}) classifier.
We considered the class weighting as an additional parameter. We integrated this into our parameter sweep 
by means of different class subsample sizes. For a \textit{location-specific} classifier we considered sample sizes of 2500 to 15000 (in steps of 2500) for each class. For a \textit{location-agnostic} classifier we sampled 10000 to 22500 (in steps of 2500) samples from each class. 

The Pareto frontiers of $F_1$-score and mean detection time for the linear SVM and the XGBoost classifiers resulting from our random parameter sweep are depicted in Fig.~\ref{fig:res_wearing_location}.
The results of the \textit{location-agnostic} classifiers are depicted in Fig.~\ref{fig:xgb_all_pareto_frontier} and 
\ref{fig:lin_svm_all_pareto_frontier}.
We see that the linear SVM and the XGBoost classifiers averaged over all wearing locations perform very similar, i.e., the Pareto frontiers are close. But neither of both model types reaches an $F_1$-score of 
one. Furthermore, we observe a strong dependency on the wearing location. We can see that the XGBoost \textit{location-agnostic} classifiers for smart devices located 
in the trouser pocket and backpack show the best solutions concerning the $F_1$-score as well as the mean detection time. 
For the linear SVM this increase in detection performance is even more pronounced regarding the trouser pocket wearing location (cf. Fig.\ref{fig:lin_svm_all_pareto_frontier}).

The results for the \textit{location-specific} classifiers are depicted in Figs.~\ref{fig:xgb_pareto_frontier} and \ref{fig:lin_svm_pareto_frontier}. For the XGBoost classifiers we observe an improvement for the trouser pocket and backpack wearing locations. The best XGBoost classifier with a mean detection time of under half a second has an $F_1$-score of $94\%$ and a mean detection time of \SI{0.34}{\second}.
The detection results of the device mounted to the rack of the bicycle are slightly better while we do not observe any 
noticeable change for the device worn in the jacket pocket.  
We measure no improvements for the Pareto optimal solutions of the \textit{location-specific} linear SVM classifier
as can be seen in Fig.~\ref{fig:lin_svm_pareto_frontier}.
Instead, the mean detection times of the solutions on the Pareto frontiers are higher than those of the linear SVM \textit{location-agnostic} classifiers. 
Only for the device in the trouser pocket we observe an increased $F_1$-score.
While for the \textit{location-agnostic} classifiers the XGBoost and linear SVM show comparable results. 
In the case of \textit{location-specific} classifiers the XGBoost outperforms the linear SVM.

One major challenge that we observed with all classifiers is that many false positives are due to miss-classifications at the 
beginning of the waiting phase. Movements, such as getting off the bike or adjustment of pedals, produce a similar sensor pattern resulting in false positive classification.
In order to investigate the starting detection performance more closely, we additionally examined shortened starting scenes. 
We cut the scenes to \SI{7}{\second} before the labeled starting movement. This removes the remaining motion originating from the stopping movement, e.g., getting off the bike. The Pareto frontiers of the shortened scenes of the XGBoost and linear SVM  \textit{location-specific} classifiers are depicted in Figs.\ref{fig:short_xgb_all_pareto_frontier} and \ref{fig:short_lin_svm_all_pareto_frontier}. 
We observe an increased $F_1$-score, i.e., the XGBoost classifier reaches an $F_1$-score of 97.6\% at mean detection time of \SI{0.35}{\second}. 


\subsection{Feature Selection}

\begin{figure*}[ht!]
	\centering
	\begin{subfigure}[t]{0.49\linewidth}
		\includegraphics[width=0.9\columnwidth]{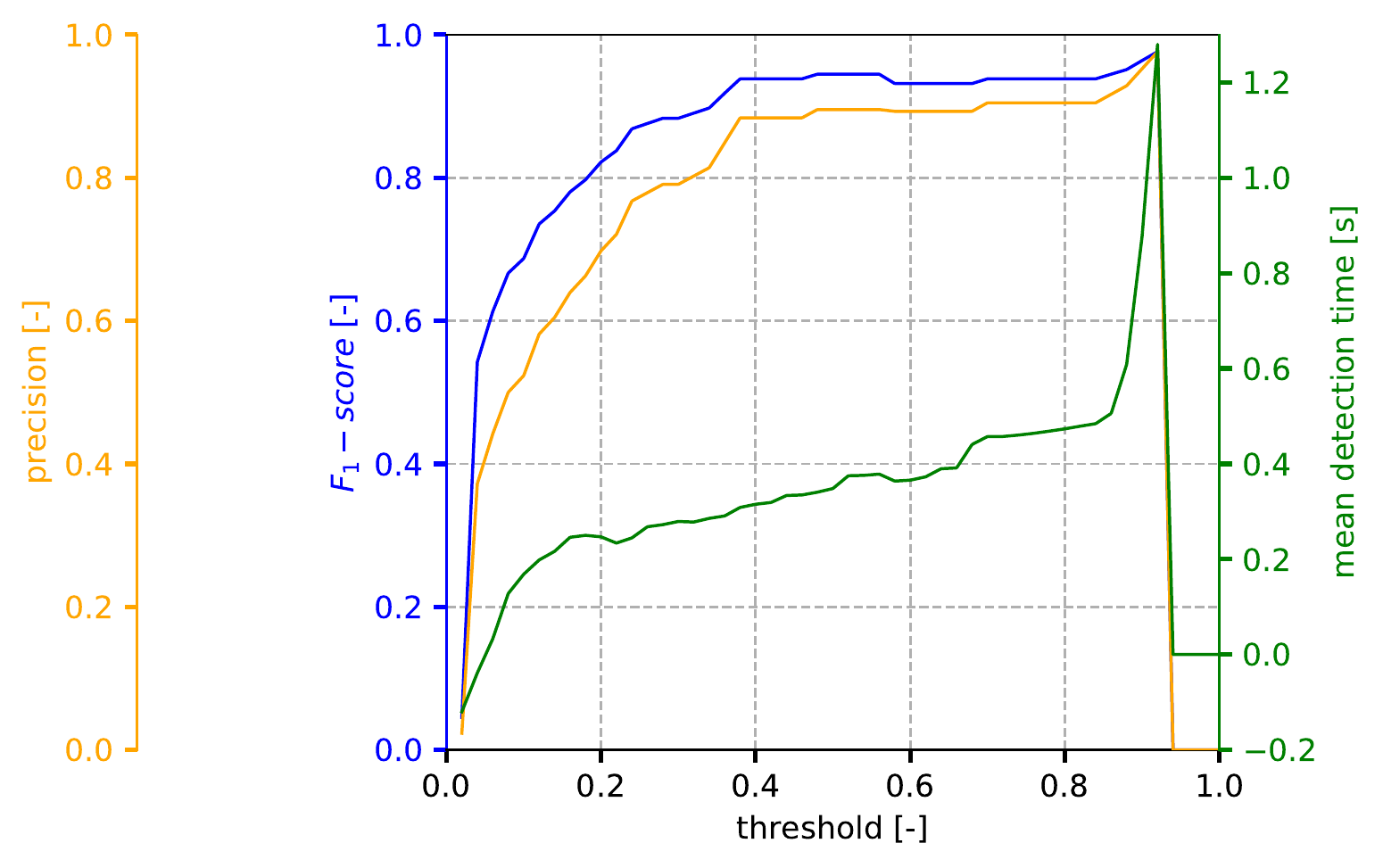}
		\caption{\textit{Location-specific} XGBoost.}
	\end{subfigure}
	\begin{subfigure}[t]{0.49\linewidth}
		\includegraphics[width=0.9\columnwidth]{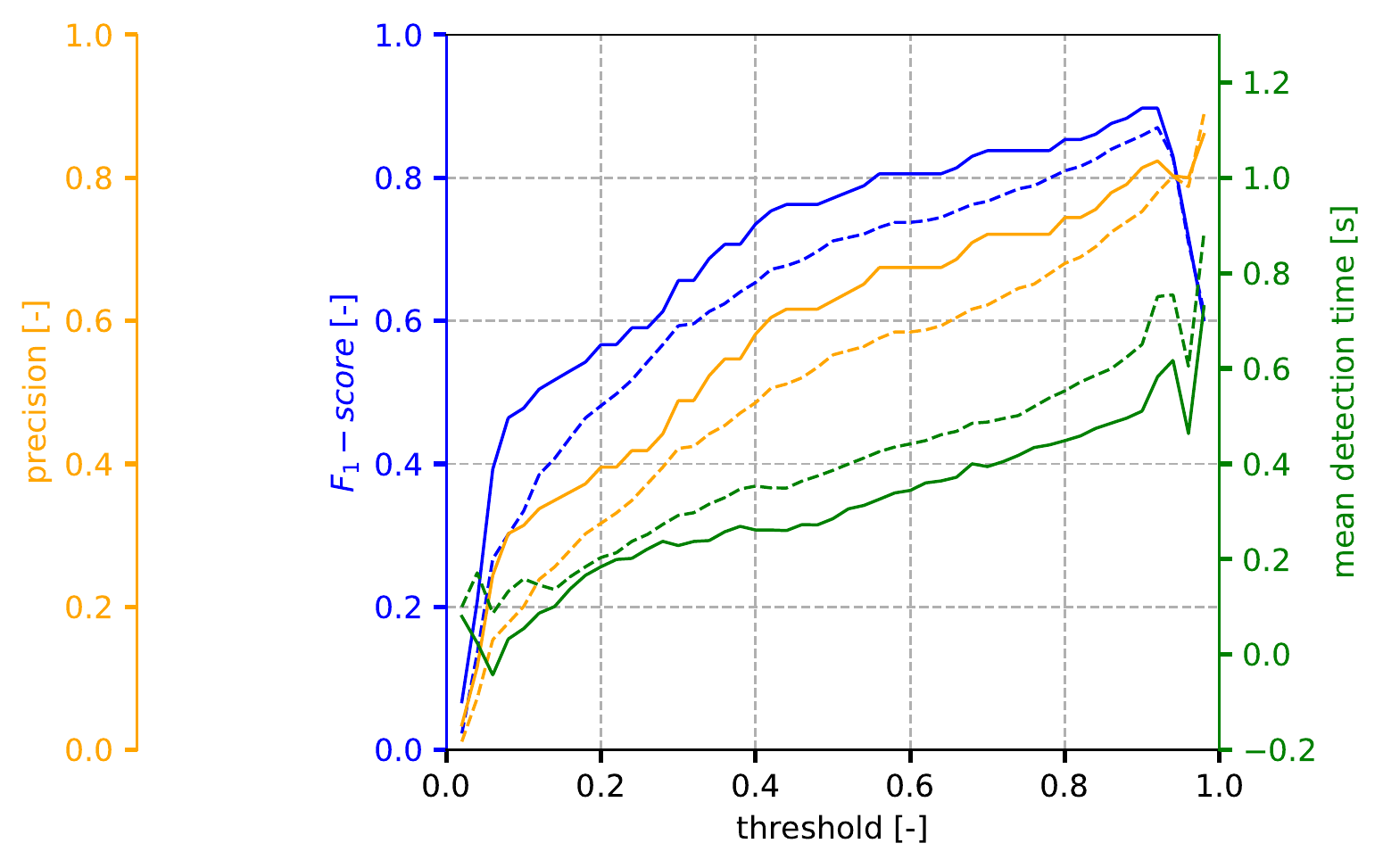}
		\caption{\textit{Location-agnostic} XGBoost.}
	\end{subfigure}
	\vskip 5mm
	\caption{$F_1$-score (blue), precision (yellow) and mean detection time (green) over the probability threshold of all starting scenes for the selected device \textit{location-specific} (left) and \textit{location-agnostic} (right) classifier evaluated for the smart device worn in the trouser pocket. The dashed lines indicates the reference evaluation involving all wearing positions.}
	\label{fig:koehler_plot}
	\vskip -2mm
\end{figure*}

\begin{figure*}[h]
	\centering
	\begin{subfigure}[t]{0.49\linewidth}
		\centering
		\includegraphics[width=0.8\linewidth]{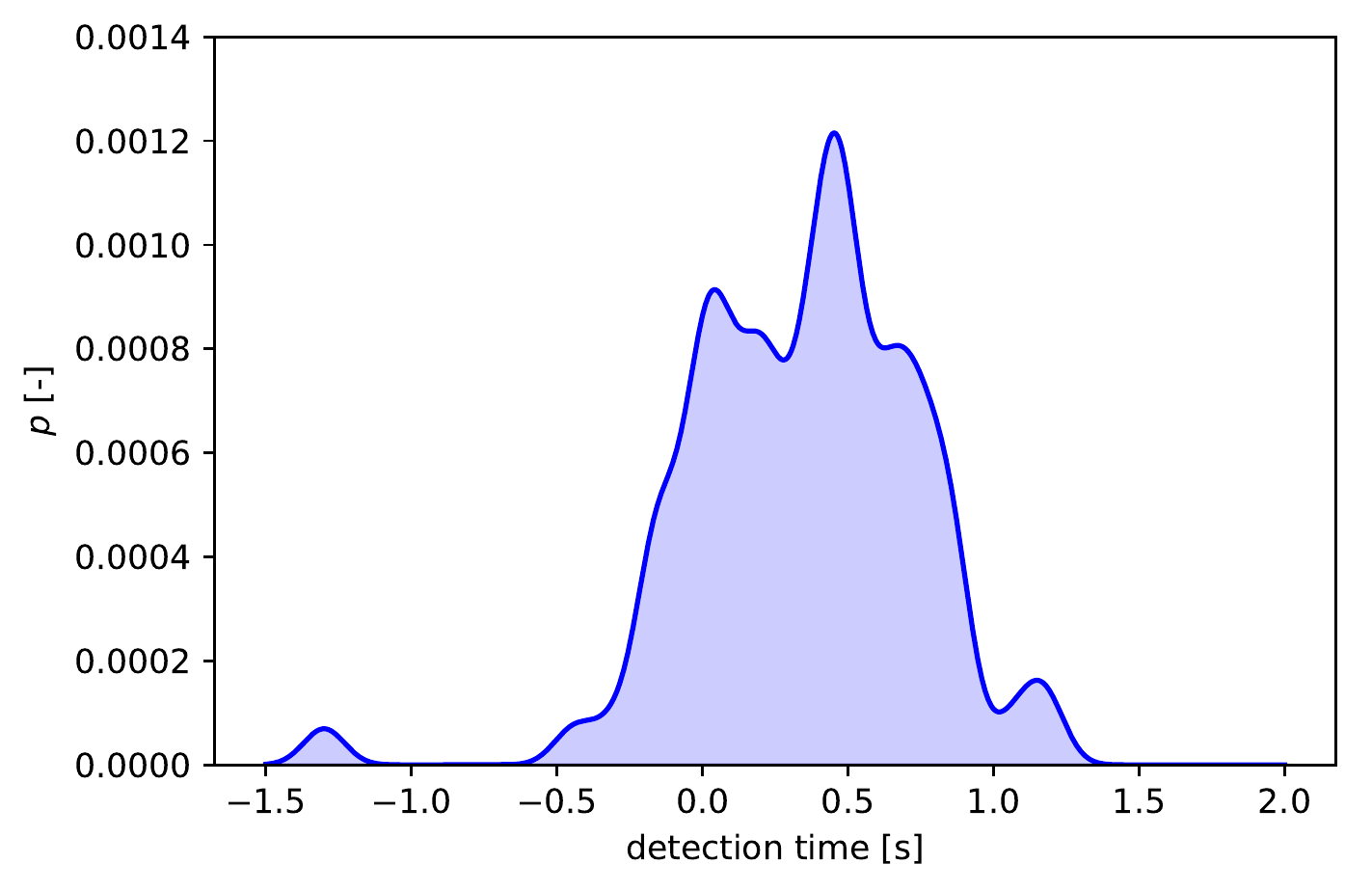}
		\vskip 2mm
		\caption{\textit{Location-specific} XGBoost.}
		\label{fig:location_specific_delay_histogram}
	\end{subfigure}
	\begin{subfigure}[t]{0.49\linewidth}
		\centering
		\includegraphics[width=0.8\linewidth]{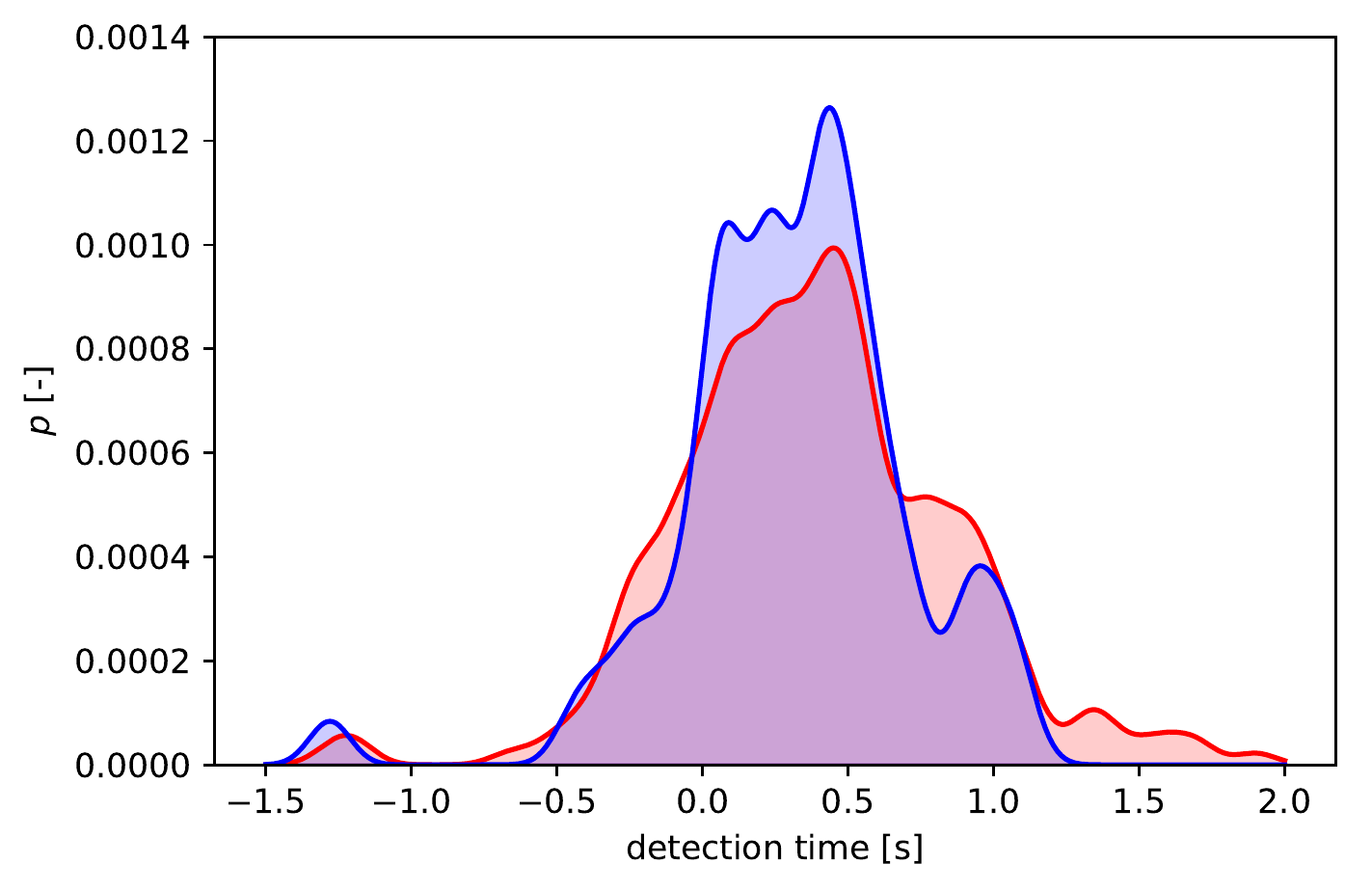}
		\vskip 2mm
		\caption{\textit{Location-agnostic} XGBoost.}
		\label{fig:location_agnostic_delay_histogram}
	\end{subfigure}
	\vskip 5mm
	\caption{Histograms of mean detection times. The histograms are smoothed by means of a kernel density estimation with a Gaussian kernel of bandwidth 70 (Blue: trouser pocket, Red: all wearing positions).
	}
	\label{fig:delay_histogram}
	\vskip -2mm
\end{figure*}

In this section, we evaluate the two-stage feature selection performed for the distinct XGBoost classifiers trained 
for different wearing locations.
 Moreover, an interpretation of the selected features with respect 
to the wearing location is given. Since we performed a ten-fold cross-validation, we considered the 
number of times a particular feature is selected as evaluation criteria. 
For the ease of understandability, we grouped different window sizes of the same features in our evaluation.
For the smart devices located in the front trouser pocket only a single feature was selected, i.e., 
the gyroscope's energy in the ground plane (with varying window sizes in different folds). 
It captures the pedalling and push off movement, which happens in the early starting phase, 
very well. The most relevant feature for the smart device located at the bicycle's rack 
is the linear accelerometer's energy in the ground plane. This feature captures the energy of the movement in driving direction. 
For the device located in the jacket pocket at the chest the most relevant features are the gyroscope's energy in the ground plane and the linear accelerometer's energy along the z-axis. The first feature captures the pedalling movement while the second feature additionally measures the downward movement of the upper body just before starting~\cite{mbHZD+17} as well as the pushing away motion to start.
The classifiers based on the smart device in the backpack use the largest number of features. 
The most important features of those are the linear accelerometer's minimum, energy, and the second polynomial degree in the ground plane as well as the gyroscope's energy, minimum, and maximum in the ground plane. 
The classifiers that are based on data originating from all wearing locations are heavily based on the energy of the linear accelerometer along the z-axis, the gyroscope's energy along the z-axis, and the residual of the linear accelerometer's DFT polynomial approximation along the z-axis.
It is noticeable that this classifier mainly uses features which are based on the z-axis. 
Many typical starting movements (e.g., pedalling, pushing away motion) are  well capture by 
features computed for the accelerometers z-axis. The features based on the gyroscopes z-axis are best 
explained by unsteady, swaying bicycle movements at starting.

\subsection{Starting Movement Detection}

We focus on the detailed evaluation of two selected starting movement detectors. 
Moreover, we show the influence of the auxiliary class used to train the classifiers.
For evaluation, we compare a \textit{location-agnostic} with a \textit{location-specific} classifier.
We restrict ourselves to XGBoost based detectors since they outperform linear SVM based detectors.
The detectors are chosen from the Pareto frontiers (cf. Fig.~\ref{fig:res_wearing_location}). As selection criteria, we require a mean detection time of at most \SI{0.4}{\second}.
This is motivated by the scenario mentioned in the introduction. But here, we additionally consider communication 
delay of modern Car2X-communication system (e.g., IEEE 802.11p) of approximately \SI{0.1}{\second}.
For the \textit{location-agnostic} classifier, we select the XGBoost classifier with an $F_1$-score of 67\% and a mean detection time of \SI{0.33}{\second}. The number of trees is 100, the maximal tree depth is 8 and the learning rate is 0.18.
 We compare this detector with the \textit{location-specific} XGBoost classifier for devices worn in the trouser pocket ($F_1$-score of $94\%$ and mean detection time of \SI{0.34}{\second}).  The number of trees is 300, the maximal tree depth is 3 and the learning rate is 0.11.
The $F_1$-score, precision and mean detection times for different probability thresholds on the \textit{moving} class are depicted in 
Fig.\ref{fig:koehler_plot}. 
We see that the optimal probability threshold for the \textit{location-specific} classifier is around 0.5.
For the \textit{location-agnostic} classifier, the $F_1$-score can be improved by applying a higher probability threshold. 
We observe that the $F_1$-score scales approximately linear with the 
mean detection time. We also see that an $F_1$-score of one is not reached, not even when the probability threshold is close to 
one. Far from it, when approaching a threshold of one the $F_1$-score deteriorates for both classifiers, i.e., 
starting is no longer detected.
Since the classifiers are calibrated, it can be concluded that the \textit{waiting} and \textit{moving} classes
cannot be entirely separated based on the selected features, i.e., the classes posses a particular level of impurity. 
Because the \textit{location-specific} XGBoost classifier for the trouser pocket uses only a single feature, this 
effect gets even more pronounced. For the \textit{location-agnostic} classifier it is more moderate.

In the following, we investigate the detection times. We examined the distribution of detection times of the shortened starting 
scenes. The histograms of detection times for the \textit{location-specific} and \textit{location-agnostic} XGBoost classifier are depicted in Figs.~\ref{fig:location_specific_delay_histogram}~and~\ref{fig:location_agnostic_delay_histogram}. To obtain a probability distribution, the histograms are smoothed with a kernel density estimation using a Gaussian kernel.
We see that for both, the \textit{location-specific} and \textit{location-agnostic} XGBoost classifiers, some starting movements 
are detected more than \SI{1}{\second} before the first movement of the bicycle wheel. The remaining detection times can be approximately described by a Gaussian centered at the mean detection time of the respective detectors. Whereas the distribution of the \textit{location-specific} classifier is narrower and the one of the \textit{location-agnostic} classifier is heavy-tailed.

Moreover, we also investigate the effectiveness of additional output smoothing. 
We found that the effects are negligible, i.e., the $F_1$-score but also the detection time is partly increased. 
We examine it as a way to fine-tune the detectors.

\begin{figure}[H]
	\vskip -4mm
	\begin{subfigure}{\linewidth}
		\centering
		\includegraphics[width=0.9\columnwidth]{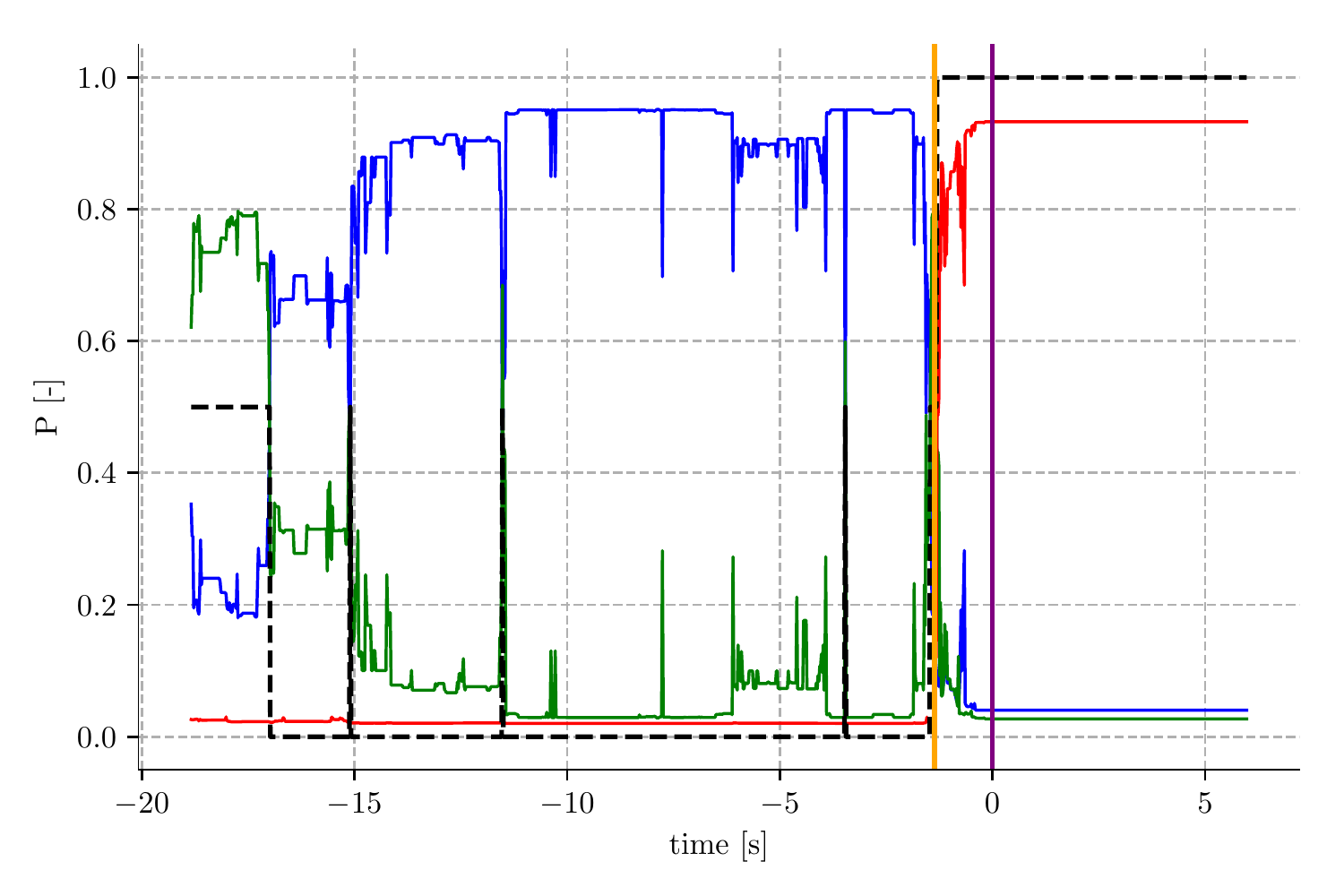}
		\vskip -3mm
		\caption{Starting movement with negative mean detection time.}
		\label{fig:pred_dv1_131}
	\end{subfigure}
	\begin{subfigure}{\linewidth}
		\centering
		\includegraphics[width=0.9\columnwidth]{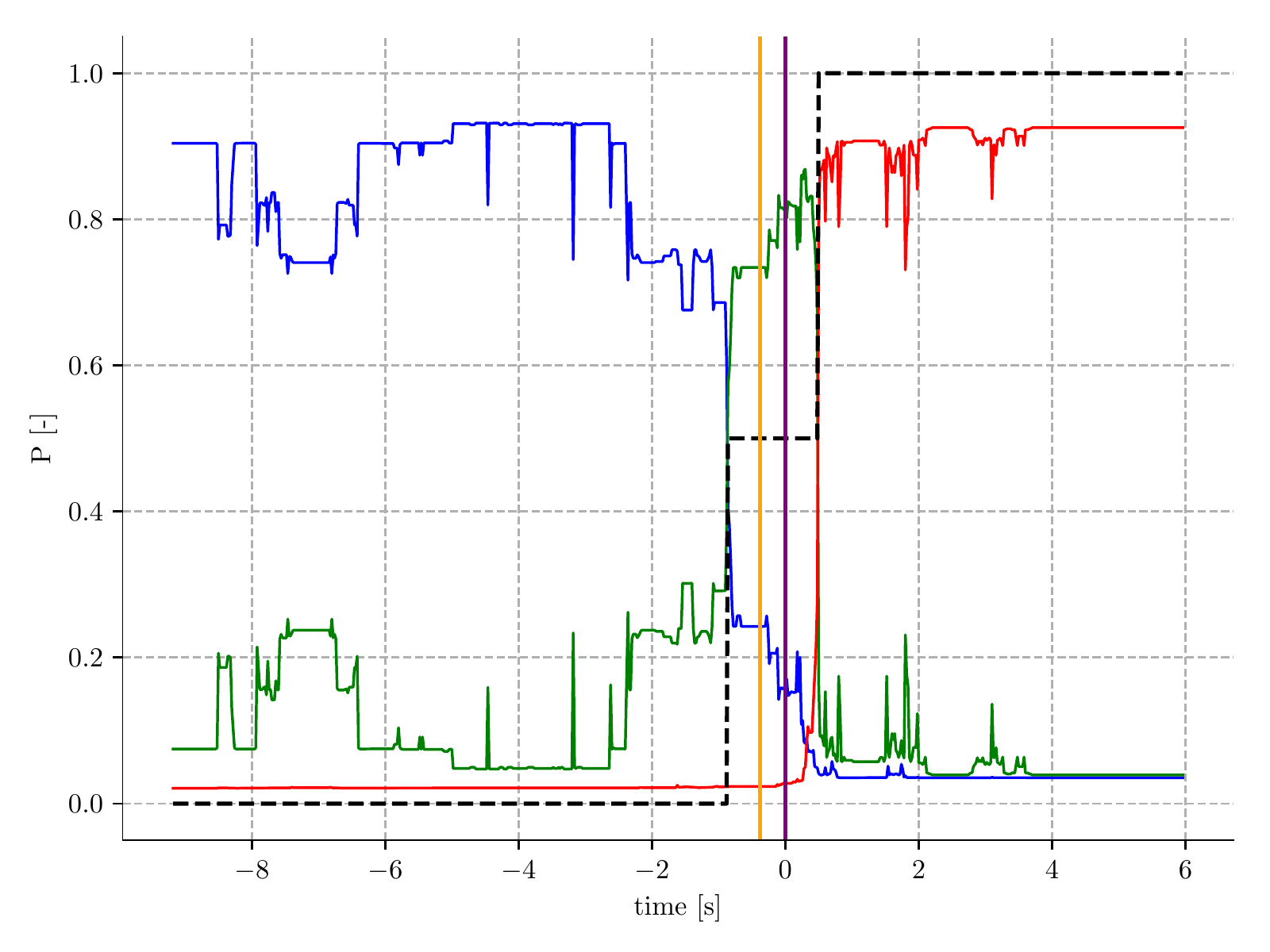}
		\vskip -3mm
		\caption{Starting movement detection with delayed detection.}
		\label{fig:pred_dv1_121}
	\end{subfigure}
	\begin{subfigure}{\linewidth}
		\centering
		\includegraphics[width=0.9\linewidth]{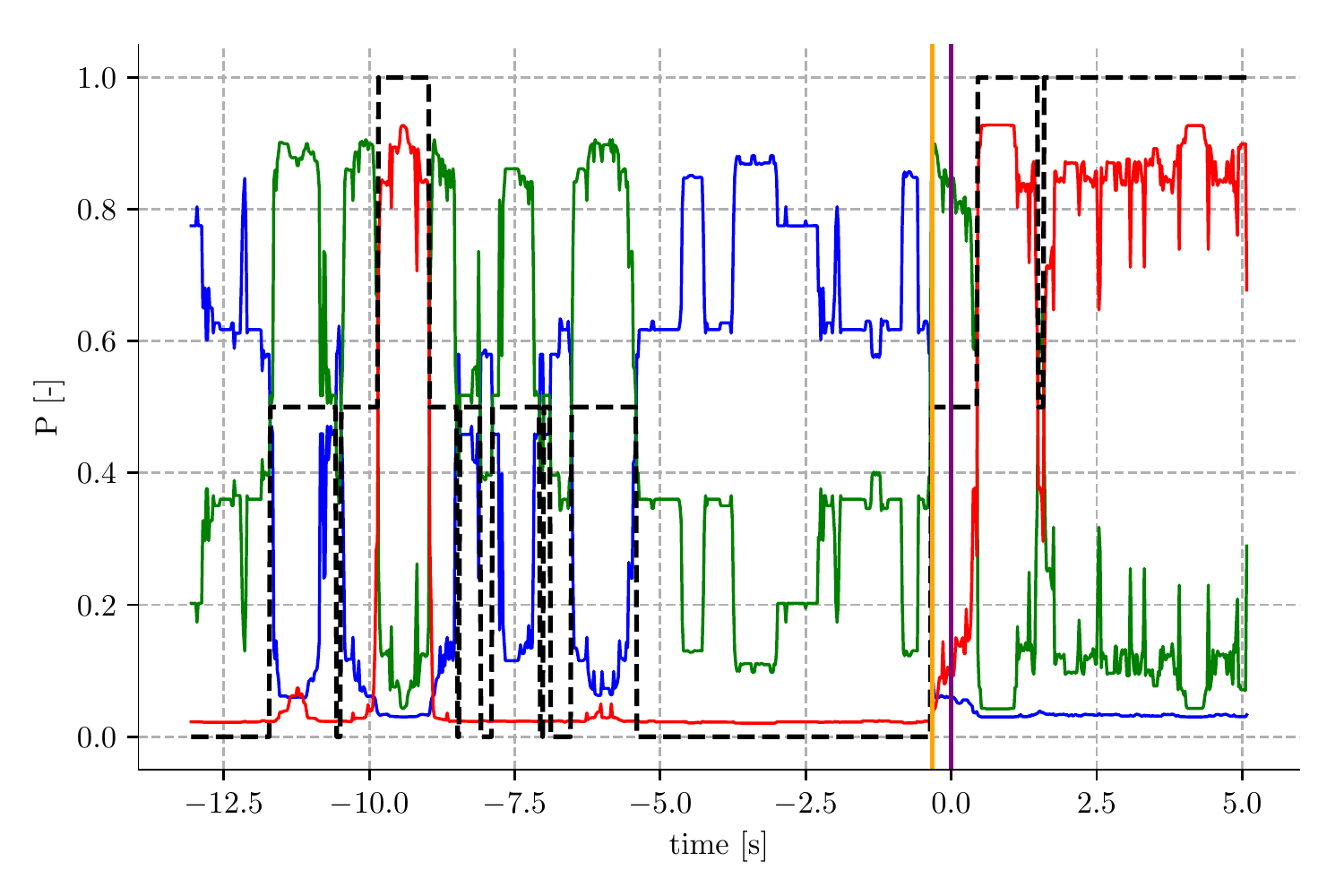}
		\vskip -3mm
		\caption{False positive starting movement detection.}
		\label{fig:pred_dv1_140}
	\end{subfigure}
	\begin{subfigure}{\linewidth}
		\centering
		\includegraphics[width=0.9\columnwidth]{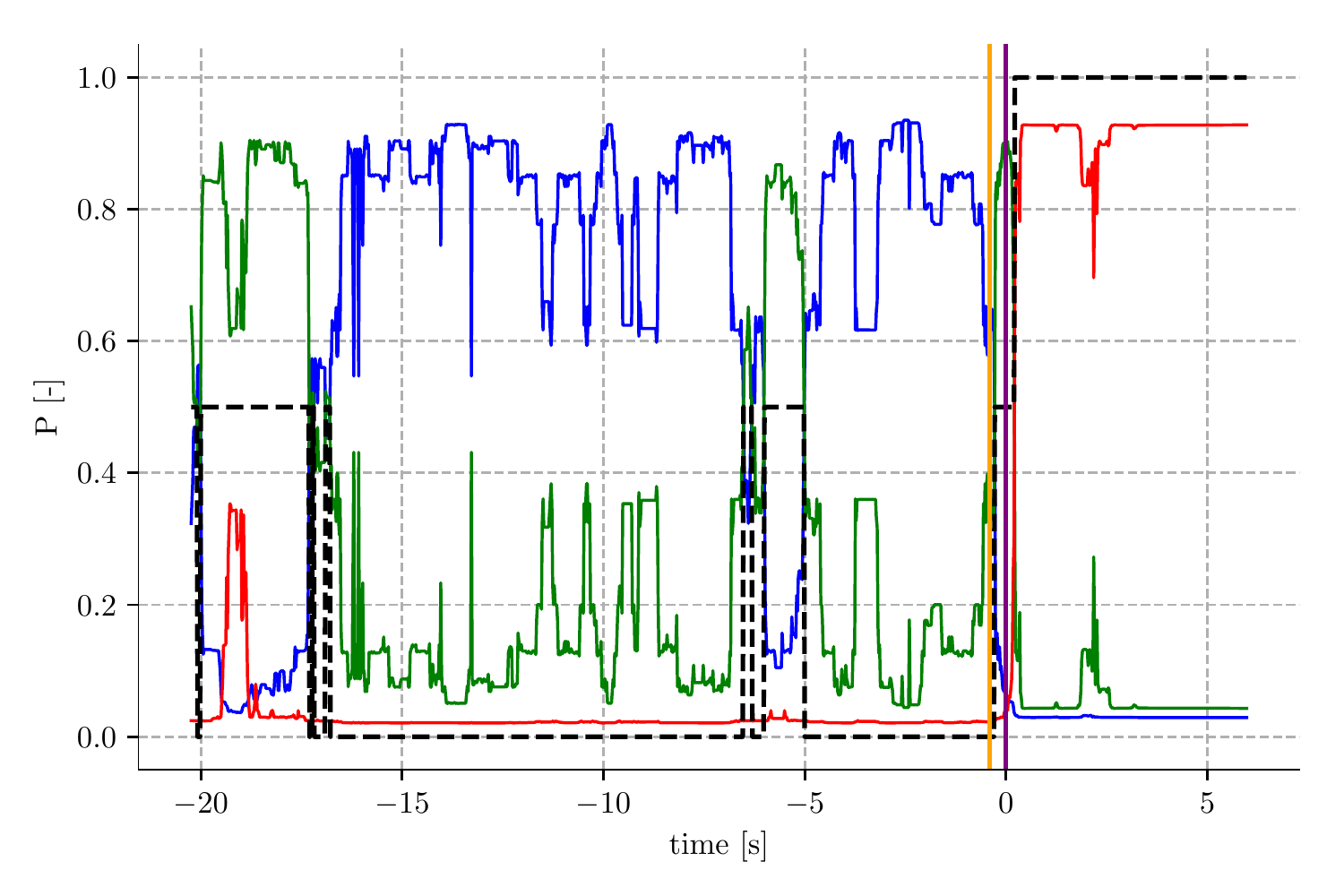}
		\vskip -3mm
		\caption{Usage of auxiliary class to avoid false positive detections.}
		\label{fig:pred_dv1_105}
	\end{subfigure}
	\vskip 4mm
	\caption{\textit{Location-specific} XGBoost detection results of selected starting scenes evaluated for smart device worn in the trouser pocket. Blue, green, and red denote the \textit{waiting}, \textit{starting}, and \textit{moving} probability. The dashed black line 
		indicates the selected class. The vertical lines denote the beginning of labeled \textit{starting} (orange) and \textit{moving} (purple) class.}
	\label{fig:selected_sample_scenes}
\end{figure}

Figure~\ref{fig:selected_sample_scenes} shows the detection results of 
four selected sample starting scenes. It depicts the results 
of the \textit{location-specific} XGBoost classifier evaluated for the device worn in the trouser pocket. In Fig.~\ref{fig:pred_dv1_131} 
a scene with a negative starting detection time is shown.
The starting movement is detected \SI{1.3}{\second} before 
the first movement of the bicycle wheel.
A starting movement with a delayed detection is depicted in 
Fig.~\ref{fig:pred_dv1_121}. We notice that the \textit{starting}
phase is detected even before the labeled \textit{starting} movement.

Regarding the \textit{moving} probability, we observe a sharp transition from the \textit{starting} to the \textit{moving} phase.
Ego-movement, e.g., adjustment of pedals or swaying from one side to the other, can lead to false positive starting detections. 
Figure.~\ref{fig:pred_dv1_140} shows such a scene containing a false positive detection. Figure.~\ref{fig:pred_dv1_105} shows the usage of the auxiliary \textit{starting} class to avoid false positive detections. It acts as a buffer zone between the \textit{waiting} and \textit{moving} classes. 

\section{\large Conclusions and Future Work}
\label{sec_conclusion}
\addtolength{\textheight}{-0.0cm}  
In this article, we presented an approach to detect the starting movement of cyclists using smart devices.
The approach is based on HAR and a modelling of the starting movement detection as a classification problem.
We introduced an additional auxiliary class modelling the transition between the \textit{waiting} and \textit{moving} classes.
This class allows to integrate early movement indicators, i.e., body movements indicating future behaviour. 
In this way we improve the robustness and reduce the detection time of the classifiers. 
It is complemented by a novel two-stage feature selection procedure, selecting robust and understandable features.

In our experiment involving 49 test subjects conducted in real-world traffic, we found  
that our detector based on the \textit{location-agnostic} XGBoost classifier reaches 
an $F_1$-score of 67\% at a mean detection time of \SI{0.33}{\second} evaluated for all wearing locations.
Although, our approach is not able to reach an $F_1$-score of one, our analysis of the detection times 
showed that a considerable amount of starting movements are detected 
before the first movement of the wheel.
We also showed that the device worn in the trouser pocket provides the best detection results.
Moreover, we demonstrated that training distinct classifiers for specific locations  
can further improve detection results, i.e., reaching an $F_1$-score of 94\% with a mean detection time of \SI{0.34}{\second} for the device worn in the trouser pocket.
Our proposed approach is still not fully competitive with
accurate video-based methods (e.g., ~\cite{ZKS+18}) as available in infrastructure and vehicles, yet.
Nevertheless, these methods fail in case of bad visibility or occlusion.
We see the information delivered by smart devices as complementing the cooperative intention detection 
process~\cite{decoint2BZH+18}, e.g., allowing to resolve occlusion situations. 

For future work, we will focus on extending our presented approach towards a fully smart device based intention detection system including improved localization abilities and trajectory forecasts.
We will generalize our approach to different VRUs such as pedestrians.
Moreover, instead of using only a single smart device, we assume that in near future VRUs will
have various wearable devices at different wearing locations, e.g., smart
watches at their wrist, sensor-equipped helmets and shoes. 
We will also investigate the continuous on-line improvement of the detection models.
Therefore, we aim to integrate imprecise labels by means of active and semi-supervised learning techniques to reduce the dependency on the high quality camera system.
Furthermore, we will investigate a more tight integration into a cooperative 
setting, i.e., combining information from smart devices and other road users for advanced VRU safety.
We will investigate opportunities to fuse this information to improve our starting movement detection approach.


\section*{\large Acknowledgment}

This work results from the project DeCoInt$^2$, supported by the German Research Foundation (DFG) within the priority program SPP 1835: "Kooperativ interagierende Automobile", grant number SI 674/11-1.


\bibliographystyle{IEEEtran}
%

{\small
\bibliography{IEEEabrv,mb}
}

\end{document}